\pdfoutput=1

\documentclass[11pt]{article}

\usepackage[]{acl}

\usepackage{times}
\usepackage{latexsym}

\usepackage[T1]{fontenc}

\usepackage[utf8]{inputenc}

\usepackage{microtype}

\usepackage{inconsolata}

\usepackage{graphicx}
\usepackage{adjustbox}
\usepackage{csquotes}
\usepackage{xcolor,color}
\usepackage{multirow}
\usepackage{enumitem}
\usepackage{multirow}
\usepackage{amsmath}
\usepackage{algorithmic}
\usepackage{algorithm}
\usepackage{pifont}
\usepackage{subcaption}
\usepackage{amssymb}
\usepackage{pifont}
\usepackage{graphicx} 
\usepackage{subcaption} 
\usepackage{array} 
\usepackage{arydshln}
\usepackage{tcolorbox}
\usepackage{fancyhdr}

\title{\textit{We Argue to Agree}: Towards Personality-Driven Argumentation-Based Negotiation Dialogue Systems for Tourism}
\author{Priyanshu Priya, Saurav Dudhate\thanks{Equal contribution}, Desai Vishesh Yasheshbhai$^*$, Asif Ekbal  \\
Department of Computer Science and Engineering, Indian Institute of Technology Patna, India  \\
        \texttt{\{priyanshu\_2021cs26, saurav\_2001cs62, desai\_2211ai07,asif}\}{\tt @iitp.ac.in}}

\begin{document}
\maketitle
\begin{abstract}
Integrating argumentation mechanisms into negotiation dialogue systems improves conflict resolution through exchanges of arguments and critiques. Moreover, incorporating personality attributes enhances adaptability by aligning interactions with individuals' preferences and styles. To advance these capabilities in negotiation dialogue systems, we propose a novel Personality-driven Argumentation-based Negotiation Dialogue Generation (PAN-DG) task. To support this task, we introduce \texttt{\textbf{PACT}}, a dataset of \textbf{P}ersonality-driven \textbf{A}rgumentation-based negotiation \textbf{C}onversations for \textbf{T}ourism sector. This dataset, generated using Large Language Models (LLMs), features three distinct personality profiles, \textit{viz.} \textit{Argumentation Profile}, \textit{Preference Profile}, and \textit{Buying Style Profile} to simulate a variety of negotiation scenarios involving diverse personalities. Thorough automatic and manual evaluations indicate that the dataset comprises high-quality dialogues. Further, we conduct comparative experiments between pre-trained and fine-tuned LLMs for the PAN-DG task. Multi-dimensional evaluation demonstrates that the fine-tuned LLMs effectively generate personality-driven rational responses during negotiations. This underscores the effectiveness of \texttt{\textbf{PACT}} in enhancing personalization and reasoning capabilities in negotiation dialogue systems, thereby establishing a foundation for future research in this domain\footnote{Resources can be accessed at \url{https://www.iitp.ac.in/\~ai-nlp-ml/resources.html}; \url{https://github.com/priyanshu-profile/PACT/}.}.

\end{abstract}

\section{Introduction}

\begin{figure}[hbt!]
\centering
    \includegraphics[width = \linewidth, scale=1.0]{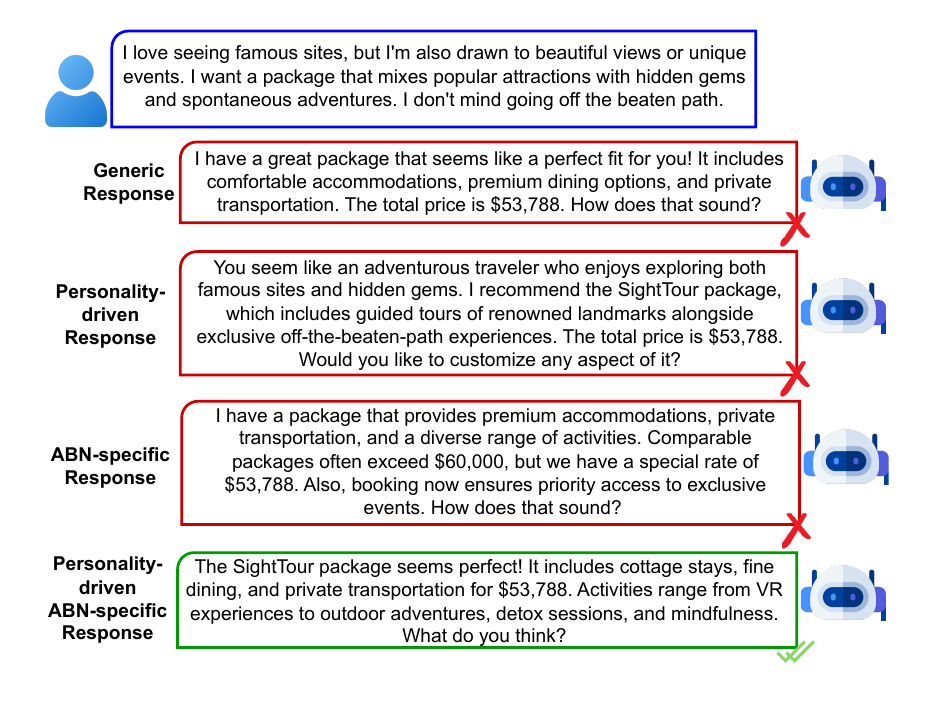}
    \caption{A conversation snippet showing how personality-driven, argumentation-based negotiation aligns with user preferences and employs ABN-specific act \textit{justify-price} to improve the negotiation outcome.}
    \label{motivation}
\end{figure}
The rapid growth of the tourism industry \cite{economic2024wttc} demands advanced negotiation dialogue systems to streamline trip booking and enhance traveler experiences. A key challenge lies in effective planning and decision-making, where LLM-based agents can contribute significantly. Unlike simple transactional tasks, tourism negotiations involve diverse user preferences, conflicting constraints, and planning across interdependent variables (e.g., price, accommodation, destination), areas where current LLMs often fall short \cite{valmeekam2023planning,gundawar2024robust}. Argumentation-Based Negotiation (ABN) \cite{rahwan2003argumentation,karunatillake2006argumentation} addresses these complexities by enabling the exchange of meta-information like justifications and persuasive arguments \cite{dimopoulos2014advances}. The tourism domain, with its complex and personalized decisions, serves as an ideal testbed for evaluating LLMs and ABN’s effectiveness in achieving collaborative outcomes. A robust ABN dialogue system for tourism enriches insights into the negotiating parties’ perspectives, ultimately fostering a more informed, transparent, and collaborative dialogue.

Personality is a crucial determinant in shaping negotiation outcomes as it significantly influences actions and behaviors of the negotiating parties \cite{sharma2013role}. Integrating personality information into ABN systems further amplifies their potential to secure collaborative outcomes by tailoring arguments and justifications to align with the personal values and negotiation styles of the parties involved, thereby fostering mutual understanding and agreement. Such personality-driven collaborative ABN dialogues enable smoother, more engaging, and highly personalized interactions, which are essential for maintaining a positive negotiating environment. For instance, as illustrated in Figure \ref{motivation}, a generic response lacks both personalization and justification, thus can lead to traveler dissatisfaction and potential negotiation failure. Personality-driven responses reflect traveler preferences but lack persuasive support, while ABN-specific responses provide justification but remain impersonal. In contrast, personality-driven ABN-specific responses integrate both personalization and persuasive reasoning (justify-price), leading to more engaging and effective negotiations. 

Motivated by these considerations, we introduce a novel \textbf{P}ersonality-driven \textbf{A}rgumentation-based \textbf{N}egotiation \textbf{D}ialogue \textbf{G}eneration (PAN-DG) task aimed at generating negotiation dialogues tailored to individuals' personalities while integrating authoritative argumentation techniques for personalized, rational, and collaborative negotiation outcomes. The PAN-DG task is structured to reflect real-life personalized negotiation scenarios and is divided into three core sub-tasks: (i) \textit{Personality Recognition}, (ii) \textit{Dialog Act Prediction}, and (iii) \textit{Response Generation}. These sub-tasks collectively address key phases of negotiation, from identifying negotiators' personalities, predicting next dialog act, to generating responses that align with these personalities and dialog acts.

To support the PAN-DG task, we present a novel dataset of \textbf{P}ersonality-driven \textbf{A}rgumentation-based \textbf{N}egotiation \textbf{C}onversations for \textbf{T}ourism, named \texttt{\textbf{PACT}}. Recognizing the labor and time-intensive nature of manually creating high-quality personalized conversational datasets \cite{zhang2018personalizing,zhong2020towards,wu2020getting,jang2022call}, \texttt{\textbf{PACT}} is automatically generated using LLMs, with human oversight through a three-stage pipeline: (1) \texttt{\textbf{PACT}} Generation, where the LLM \texttt{Gemini-1.5} \cite{team2023gemini} is prompted to generate negotiation dialogues tailored to diverse personalities. Considering the multifaceted nature of individual's personality \cite{bayram2017decision}, this work delineates three distinct categories of personality profiles: \textit{Argumentation Profile} that elucidates how individuals negotiate and argue, \textit{Preference Profile} that encapsulates their trip preferences, and \textit{Buying Style Profile} that characterizes their purchasing behaviors; (2) \texttt{\textbf{PACT}} Filtering, which employs a \texttt{GPT-4}-based \cite{openai2024chatgpt} mixture-of-experts approach to ensure high-quality dialogues; and (3) \texttt{\textbf{PACT}} Quality Assessment, involving automated and manual evaluations to assess the dialogue quality. This pipeline results in a first-of-its-kind dataset, comprising 8K+ personality-driven argumentation-based negotiation dialogues for tourism. 

Using this dataset, we conduct comprehensive comparative experiments involving pre-trained and fine-tuned LLMs for aforementioned sub-tasks, establishing strong baselines for future investigation. Our dataset construction approach and task methodology are readily extensible to various LLMs, tasks, languages, and domains, offering a valuable tool for addressing challenges in data synthesis, eventually facilitating the development of more robust and scalable ABN dialogue systems.

In summary, our contributions are: (i) Introduce a novel personality-driven argumentation-based negotiation dialogue generation task to generate dialogues aligned with negotiators' distinct personalities; (ii) Develop an efficient multi-stage dataset generation pipeline using LLMs with human supervision to automate dataset creation; (iii) Curate a new dataset, \texttt{\textbf{PACT}}, comprising 8K+ personality-driven argumentation-based negotiation conversations focused on tourism domain; (iv) Define a rich set of three personality profiles: \textit{Argumentation Profile}, \textit{Preference Profile}, and \textit{Buying Style Profile}, capturing diverse negotiation behaviors; (v) Establish strong LLM-based baselines for three sub-tasks: personality recognition, dialog act prediction, and response generation, emulating real-life personalized negotiation scenarios.

\section{Related Work}
\textbf{Negotiation Dialogue Systems.} In recent years, negotiation dialogue systems have garnered significant attention \cite{zhan2024let,priya2025genteel}. Early studies focused on strategic modeling to facilitate negotiation \cite{lewis2017deal} and used dialogue acts to track negotiators' behavior \cite{he2018decoupling}. Strategic negotiation frameworks have also been applied to specific scenarios, including campsite item negotiations \cite{chawla-etal-2021-casino}, job offers \cite{yamaguchi2021dialogue}, and electronic items \cite{ahmad2023ina}. Lately, LLMs have been employed in negotiation tasks, enabling more sophisticated strategies and interactions. Balloon price negotiations were explored using LLM-based agents \cite{fu2023improving}, and a plug-and-play policy planner was introduced to optimize strategy formulation in LLMs \cite{deng2023plug}. Strategic reasoning has also been enhanced through chain-of-thought (CoT) prompting \cite{gandhi2023strategic}, which also demonstrated success in facilitating multi-agent agreements during interactive games \cite{abdelnabi2023llm}.

While these studies primarily model negotiation through the lens of game theory and action selection \cite{nash1950bargaining,baarslag2013evaluating}, our research takes a novel approach by adopting an argumentation-based framework for collaborative negotiation within the tourism domain. To our knowledge, this work pioneers the exploration of argumentation-based negotiation dialogue systems.

\textbf{Personality-driven Dialogue Generation.} Personalization is critical in enhancing the effectiveness of dialogue systems \cite{weston2018retrieve,zheng2019personalized,zheng2020pre}. Early research focused on utilizing user profiles to generate tailored responses across various applications like chit-chat \cite{firdaus2022enjoy}, persuasive communication \cite{tiwari2023towards}, therapy \cite{mishra2023therapist}, and healthcare \cite{dou2023plugmed,liu2024context}. However, research on personalization in negotiation dialogues is relatively sparse. Drawing on psychological evidence that personality \cite{sharma2013role} significantly influences negotiation dynamics, researchers estimated opponents' personalities to improve negotiation outcomes \cite{zhang2020learning,yang2020improving} and proposed ranking their preferences to achieve successful negotiations \cite{chawla2022opponent}. \citet{priya2024trip} leverages travel persona information to generate personalized responses during negotiation. Other research provided empirical insights into strategic impact of Big-Five personality traits on negotiation outcomes \cite{huang2024personality,noh2024llms}. Strategic planning in negotiations has been improved by incorporating diverse simulators and integrating psychological factors, such as Big-Five personality traits, and decision-making styles, leading to a deeper understanding of human-like negotiation dynamics \cite{zhang2024strength}.

While existing studies primarily focus on Big-Five personality traits and opponent modeling in negotiation, our research adopts a more comprehensive approach by defining three distinct personality profiles - Argumentation, Preference, and Buying Style to broaden the application of personality-driven dialogue generation models in negotiation.

\textbf{Dialogue Generation using LLMs.} Recent research highlights the efficacy of LLMs in dialogue generation \cite{kim2022botstalk,abdullin2023synthetic,soudani2024survey}. Some studies utilized expert-crafted dialogues as in-context examples to guide LLMs in synthesizing social conversations \cite{abdullin2023synthetic,chen2023places}. However, the manual creation of these dialogues is labor-intensive and can lead to inconsistencies in quality. To mitigate this, other works \cite{lee2022personachatgen,jandaghi2023faithful,han2024psydial} proposed a framework for generating persona-based chit-chat dialogues without manual intervention. Building on these efforts, we introduce a novel LLM-based multi-stage pipeline that generates personality-driven, argumentation-based negotiation conversations and incorporates a robust filtering process complemented by comprehensive quality assessments to ensure quality dialogues.

\begin{figure*}[hbt!]
\centering
    \includegraphics[width = 0.9\linewidth, scale=1.0]{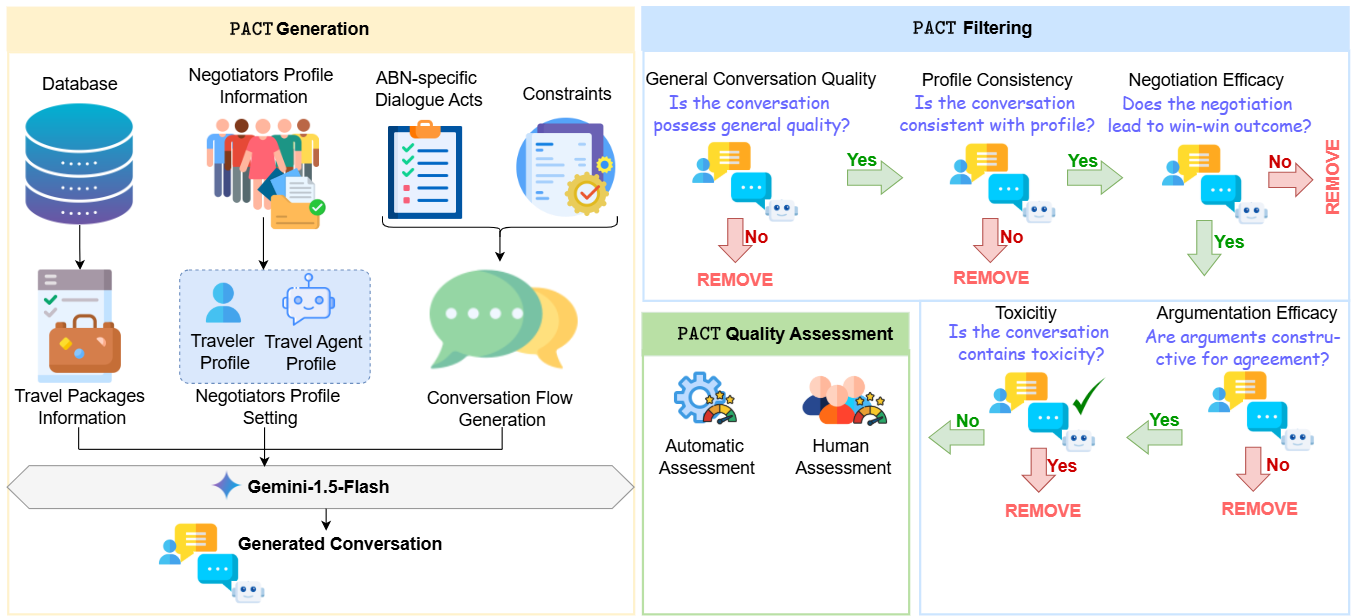}
   \caption{The overall pipeline of \texttt{\textbf{PACT}} dataset construction.} 
    \label{dataset_creation_pipeline}
\end{figure*}

\section{\texttt{PACT} Dataset Construction Pipeline}

To develop the \texttt{\textbf{PACT}} dataset with high-quality personality-driven argumentation-based negotiation conversations for tourism domain, we leverage the advanced capabilities of LLMs and design a dataset construction pipeline with three stages: (1) \texttt{\textbf{PACT}} Generation, (2) \texttt{\textbf{PACT}} Filtering, and (3) \texttt{\textbf{PACT}} Quality Assessment, as shown in Figure \ref{dataset_creation_pipeline}.

\subsection{\textbf{\texttt{PACT}} Generation}

\label{PACT_generation}
The \texttt{\textbf{PACT}} dataset consists of conversations between a travel agent and a traveler, focusing on argumentation-based negotiation (ABN) over tourism package elements such as price, destination, accommodation, etc. These dialogues include offers, justifications, counterarguments, and reasoning to reach mutually beneficial agreements, reflecting realistic negotiation dynamics. The generation process involves four key steps:

\textbf{Travel Packages Information Collection.} We compile a database of travel packages that serves as the foundation for negotiation dialogues. The package information is sourced from websites such as \href{https://www.makemytrip.com/}{MakeMyTrip}, \href{https://www.tripadvisor.com/}{TripAdvisor}, and \href{https://www.yatra.com/}{Yatra.com}, selected based on popularity, destination types, price ranges, package inclusions, and provider diversity to capture a broad spectrum of traveler preferences. This data is then used to create 10 distinct travel packages, each with a unique name, description, and specifications on amenities, services, and pricing, ensuring that dialogues reflect real-world tourism scenarios. The list of packages can be found in Appendix \ref{package}.

\textbf{Negotiators Profile Setting.} In real-world negotiation, negotiators exhibit varied personalities that influence their interactions. Experiments show that when a negotiator's personality profile is absent, LLMs generate dialogues with repetitive themes. To ensure the dialogues reflect distinct personalities and simulate realistic negotiation dynamics, each party is assigned a unique personality. These profiles dictate how they negotiate, respond to offers, and articulate their reasoning. 

While the Big-Five Personality Traits framework \cite{roccas2002big} offers a broad view of human behavior, it lacks the domain-specific granularity and may misrepresent individual differences \cite{block1995contrarian}, which is required for negotiation in the tourism context. In particular, it does not model domain-dependent factors like tourism preferences, argumentation stances, or purchasing behavior, which are essential for shaping meaningful negotiation in the tourism domain. To model these critical aspects, we introduce three domain- and task-specific profiles that directly influence different facets of negotiation behavior. For both travel agent and traveler, we define an \textit{Argumentation Profile}, and for travelers, we introduce two additional profiles: \textit{Preference Profile} and \textit{Buying Style Profile}. For both travel agent and traveler, we define an \textit{Argumentation Profile}, and for travelers, we introduce two additional profiles: \textit{Preference Profile} and \textit{Buying Style Profile}.

The \textit{Argumentation Profile} defines how a negotiator persuades, accepts, or rejects offers and arguments during negotiation, influencing their stance and responses to opposing views. In our work, we delineate four argumentation profiles based on negotiators' typologies \cite{amgoud2001agent,heras2013case}: Agreeable (Ag), Disagreeable (Di), Open-minded (Om), and Argumentative (Ar). 

During dataset generation, travelers are randomly assigned either an Ag or Di profile, while travel agents receive an Om or Ar profile. This assignment ensures a balanced and realistic negotiation environment by incorporating diverse negotiation behaviors. Travelers may either prioritize ease and cooperation (Ag) or adopt a more selective, value-driven, and uncooperative approach (Di), frequently challenging offers, expressing skepticism, and rejecting proposals unless presented with compelling justifications. In contrast, travel agents can take on a user-centric, flexible role (Om) or adopt a more assertive, aggressive, and profit-driven stance (Ar), actively pushing for higher gains and resisting concessions to maximize their advantage in negotiations. This strategic assignment decouples agent behavior from traveler characteristics, thereby preventing unintended correlations, enhancing dialogue diversity, and the dataset’s applicability across various negotiation scenarios in the tourism domain.

\textit{Preference Profile} is specific to the tourism domain and delineates travelers' trip preferences that are critical for understanding their travel behaviors. A trip comprises both essential features like price, destination, accommodation, and meals, as well as supplementary features like shopping areas or monuments. While essential features are relevant to all travelers, their importance may differ across individuals, whereas supplementary features are shaped by personal choices and priorities. To capture unique travel preferences through diverse features, we define ten preference profiles based on the travel personality framework \cite{park2010travel}: Culture Creature (CC), Action Agent (AAg), Avid Athlete (AAt), Thrill Seeker (TS), Trail Trekker (TT), Escapist (E), Shopping Shark (SSh), Boater (B), Sight Seeker (SSe), Beach Lover (BL). \textit{Buying Style Profile} outlines travelers' purchasing behaviors and motivations, influencing their engagement with agents and decision-making during negotiations. We identify three buying style profiles based on the user's purchasing behavior \cite{oumayma2023predicting}: Quality-concerned (QC), Budget-concerned (BC), and Budget-\&-Quality-concerned (B\&QC). Descriptions of these argumentation, preference, and buying style profiles can be found in Appendix \ref{profile_desc}.

\textbf{Conversation Flow Generation. }The generation of conversation flows is a multi-step process designed to capture complex dynamics of personality-driven ABN. It includes ABN-specific dialog acts formulation and ABN pathway encoding. 

\textbf{\textit{ABN-specific Dialog Acts Formulation}.} ABN conversations necessitate careful formulation of dialogue acts addressing the intricacies of argumentation, persuasion, and structured reasoning integral to negotiation. To effectively capture these nuances and represent the interactive dynamics between traveler and agent for developing a robust ABN dialogue system, we devise a set of 23 ABN-specific dialogue acts categorized into three types: (i) negotiation dialog acts, (ii) argumentation dialog acts, and (iii) general dialog acts. 

Negotiation dialogue acts capture the transactional dynamics in proposing, modifying, or accepting deal terms, including offers, counteroffers, and concessions related to price, features, or services. These acts include Negotiate-price-increase, Negotiate-price-decrease, Negotiate-price-nochange, Negotiate-add-X, and Negotiate-remove-X. Argumentation dialogue acts address reasoning, persuasion, and justifications when negotiators present arguments to support proposals or counter opposing views, aiming to convince each other. These acts include Concern-price, Disagree-price, Justify-price, Assurance-price, Disagree-X, Justify-X, and Assurance-X. General dialogue acts encompass broader interactions, such as greetings, questioning, clarifications, and so on to facilitate smooth communication. These include Greet-Ask, Inform, Elicit-preference, Ask-price, Tell-price, Ask-clarification-X, Provide-clarification-X, Provide-consent, Consent-response, Accept, and Acknowledge-acceptance. Descriptions of these dialogue acts can be found in Appendix \ref{acts_desc}.

\textbf{\textit{ABN Pathway Encoding}.} It involves systematically generating a structured sequence of dialog acts to simulate a rational and coherent negotiation pathway between a traveler and a travel agent. This process begins with the simulation of the traveler’s behavior, where their preference, buying style, and argumentation profiles are randomly selected to reflect diverse negotiation scenarios. A probability distribution over amenities determines the likelihood that the traveler accepts or rejects specific proposals, while their budget and tolerance limits are modeled using a normal distribution. These parameters define the initial constraints of the traveler and adapt dynamically based on the decisions made during the negotiation. Together, the agent's behavior is simulated by defining its argumentation profile, which is influenced by the traveler’s preferences and budget. The agent initially assumes a uniform probability distribution for traveler’s preferences and refines it based on their responses, allowing dynamic adjustments to align with traveler's preferences, such as adding or removing amenities.

To illustrate the flow, consider a typical negotiation. The conversation begins with the agent executing a `Greet-Ask' act to welcome the traveler and inquire about their preferences. The traveler responds with an `Elicit-preference' act to articulate their preferences. Subsequently, the agent uses the `Inform' act to propose a package detailing amenities, services, and costs. If the traveler raises concerns about the price (`Concern-price') or disagrees with the offer (`Disagree-price'), the agent may justify its proposal (`Justify-price') or make concessions (`Negotiate-price-decrease'). When necessary, the conversation transitions to an argumentation or negotiation phase, where both parties present counter-offers or arguments to reach mutually beneficial agreement. The negotiation concludes when an agreement is reached or the predefined turn limit is exhausted, marked by the agent’s execution of an `Accept' act. The generation of these conversation flows ensure that conversations adhere to rational progressions, closely aligning with real-world negotiation patterns. Besides, this approach facilitates scalability by enabling the adaptation of conversation flows across linguistic styles while also allowing detailed simulations of agent and traveler behaviors, effectively capturing diverse negotiation scenarios.

To further ensure rationality, this module employs specific constraints to maintain focus during negotiation. First, it monitors negotiation specifics on the ongoing deal for each travel package, including the agent’s minimum and current prices, traveler’s proposed price, and tolerance limit ($\phi$). Second, to promote mutually beneficial outcomes, price negotiations are capped at a predefined number of conversational turns. When this limit is reached, the conversation shifts to negotiation phase with acts like `Negotiate-add-X' or `Negotiate-remove-X'. If the traveler is dissatisfied or needs further discussion on the price, the conversation enters the argumentation phase. This phase is also limited to a specified number of turns based on the argumentative profiles of both parties to keep the dialogue focused and prevent excessive prolongation of negotiation process.

Third, to determine pricing in subsequent turns, we adopt a strategy that assumes a gradual reduction in price disparity, with increments for traveler and decrements for agent across turns following \citet{faratin1998negotiation}. The equations governing price adjustment between traveler ({\small $\mathcal{P}_t$}) 
and agent ({\small $\mathcal{P}_a$}) are: {\small $\mathcal{P}_{a_k}=\mathcal{P}_{t_{k-1}}+(\mathcal{P}_{a_{k-1}}-\mathcal{P}_{t_{k-1}})e^{-ck}$ and $\mathcal{P}_{t_k}=\mathcal{P}_{a_{k-1}}+(\mathcal{P}_{a_{k-1}}-\mathcal{P}_{t_{k-1}})e^{-ck}$}, where $c$ is a constant regulating the rate of price adjustments between conversational turns. A higher $c$ indicates a more aggressive concession strategy, while a lower $c$ denotes a conservative approach to price changes by the agent. In our approach, the agent is assigned a higher $c$ to reflect its flexibility in adjusting prices, whereas the traveler is given a lower $c$ to signify a stronger adherence to budget constraints. The value of $c$ for the agent is also contingent upon the initial price difference between two parties; specifically, in low-budget scenarios characterized by a significant price gap, a higher $c$ is adopted, whereas, in high-budget situations with a narrower gap, a lower $c$ is applied (c.f. Table \ref{value_c} in appendix). Throughout negotiation, agent will execute an `Accept' dialog act if traveler's proposed price is $\leq (\mathcal{P}_{a_k} - \phi * \mathcal{P}_{a_k})$, signaling the conclusion of negotiation. 

\textbf{Conversation Generation via Prompting.} Once the conversation flow is generated, the \texttt{Gemini-1.5-Flash} model (A preliminary analysis for LLM selection is given in Appendix \ref{llm_selection}) is prompted in a one-shot setting to generate conversations. Each prompt consists of a task overview and an utterance corresponding to the intended dialog act. Task overview clarifies scenario and goal of dialog act. 
The utterance corresponding to dialog act is manually generated based on task overview. Conversation generation flow chart, sample prompts, and example conversation are provided in Figure \ref{dataset_creation}, Table \ref{conv_generate_prompt}, Table \ref{sample_conversation}, respectively, in appendix.

\subsection{\texttt{PACT} Filtering}
A thorough examination of the \texttt{\textbf{PACT}} dataset reveals four distinct categories of erroneous conversations: (1) Empty utterances; (2) Repetitive utterances; (3) Insufficient interaction rounds; (4) Insufficient dialog act annotations; and (5) Improper conversation opening/closing. To enhance the dataset quality, we remove these undesirable conversations. Besides this initial post-processing, we implement a filtering mechanism to eliminate conversations that do not adhere to specified prompt conditions. Given the cumbersome nature of manual filtering, we capitalize on the evaluative capabilities of \texttt{GPT-4} \cite{liu2023g}. Specifically, we design a filtering pipeline for the \texttt{\textbf{PACT}} dataset, employing a mixture-of-experts approach based on  \texttt{GPT-4} \cite{openai2024chatgpt} to assess the multifaceted nature of the generated conversations. We devise five experts, each evaluating different facets of the conversation: (i) General Conversation Quality Expert (GCQE) to assess dialogue quality, including Coherence, Consistency, Diversity, Topic Depth, Understanding, Flexibility, and Likeability using Fine-Grained Evaluation of Dialog (FED) metrics \cite{mehri2020unsupervised}; (ii) Personality Consistency Expert (PCE) to evaluate the alignment of generated conversation with the traveler's and travel agent's personalities; (iii) Negotiation Efficacy Expert (NEE) to evaluate the effectiveness of negotiation between traveler and travel agent to ensure a win-win scenario; (iv) Argumentation Efficacy Expert (AEE) to evaluate the ability of negotiating parties to manage arguments and present persuasive rationale to achieve mutually beneficial outcomes; and (v) Toxicity Expert (TE) to detect harmful, biased, or offensive conversations.

These experts are implemented by prompting \texttt{GPT-4} in a zero-shot setting to rate the conversations on a 1-3 scale (1: No, 2: Somewhat, 3: Yes) except for Consistency, using a 0-1 scale (0: Consistent, 1: Inconsistent). Each rating is accompanied by an explanation to incite the rationale behind the assessment. The prompt templates for expert evaluation are furnished in Table \ref{filter_prompt1} and Table \ref{filter_prompt2} in appendix. The conversations that obtain a score of `3' from all experts and `1' from the Consistency and Toxicity experts are retained. Any conversation failing at any expert evaluation step is discarded from the dataset.

\subsection{\texttt{PACT} Quality Assessment}
\label{PACT_quality_assessment}

\textbf{Automatic Assessment.} Given that dialogues are generated through prompting the LLM, a significant concern arises regarding the limited lexical and semantic diversity of the synthetically generated conversations. To address this, we evaluate lexical diversity (differences in
word choice) using the BLEU-1/2 (B-1/2) \cite{papineni2002bleu}, METEOR (MET) \cite{banerjee2005meteor}, ROUGE-L (R-L) \cite{lin2004rouge}, DISTINCT-1/2 (D-1/2)  \cite{li2015diversity} and Self-BLEU-1/2 (SB-1/2) \cite{zhu2018texygen} metrics following \cite{wang2021naturalconv,chen2023places}. As reported in Table \ref{auto_res}, the generated conversations yield low scores for B-1/2, MET, R-L, and SB-1/2 coupled with high D-1/2 scores. These findings demonstrate that dialogue utterances exhibit substantial lexical diversity. 

Recognizing that utterances containing similar words may convey different meanings, while those using different words can express similar meanings \cite{yarats2018hierarchical}, we also assess semantic diversity (diversity of the content) through the Sem-Ent \cite{han2022measuring} metric. As shown in Table \ref{auto_res}, the high Sem-Ent score suggests that dialogues maintain significant semantic diversity, reflecting varied and contextually appropriate meanings across utterances. 

For evaluating the quality of text generated by LLMs, we employ FED \cite{mehri2020unsupervised} and G-Eval \cite{liu2023g} metrics to assess the quality of dialogues generated by the LLM. FED and G-Eval evaluate dialogues across 18 and 4 aspects, respectively. The list of evaluated aspects for FED and G-Eval are furnished in Appendix \ref{human_dataset}. We compute the average across all evaluated aspects to constitute the final `FED Score' and `G-Eval Score'. 
The result presented in Table \ref{auto_res} reveal high FED and G-Eval scores, indicating that the generated conversations mimic key aspects of human-to-human interactions, further validating their richness and naturalness. 

\begin{table}[!hbt]
\centering
\begin{adjustbox}{max width=\linewidth}
\begin{tabular}{l|cccccccc}
\hline
\textbf{Criteria} & \textbf{B-1/2}  & \textbf{MET} & \textbf{R-L} & \textbf{D-1/2} & \textbf{SB-1}$/$\textbf{2} & \textbf{Sem-Ent} & \begin{tabular}[c]{@{}c@{}}\textbf{FED} \\ \textbf{Score}\end{tabular} & \begin{tabular}[c]{@{}c@{}}\textbf{G-Eval} \\ \textbf{Score}\end{tabular}\\ \hline
\textbf{Value} & \begin{tabular}[c]{@{}c@{}}0.14/0.06 \\ (0.02/0.01)\end{tabular} & \begin{tabular}[c]{@{}c@{}}0.14 \\ (0.02)\end{tabular}& \begin{tabular}[c]{@{}c@{}}0.16 \\ (0.02)\end{tabular}& \begin{tabular}[c]{@{}c@{}}0.42/0.80 \\(0.07/0.06) \end{tabular} & \begin{tabular}[c]{@{}c@{}}0.07$/$0.05  \\ (0.08$/$0.10) \end{tabular} & \begin{tabular}[c]{@{}c@{}}0.85  \\ (0.02) \end{tabular}& \begin{tabular}[c]{@{}c@{}}4.08 \\ (4.56)\end{tabular} & \begin{tabular}[c]{@{}c@{}}3.76 \\ (1.23) \end{tabular} \\\hline
\end{tabular}
\end{adjustbox}
\caption{Result of automatic assessment of \texttt{\textbf{PACT}}. Values are averaged across dialogues (standard deviation in parentheses).}
\label{auto_res}

\end{table}

\begin{figure*}[hbt!]
    \centering
    \begin{subfigure}[b]{0.3\textwidth}
        \centering
        \includegraphics[width=\linewidth, trim=20pt 0pt 30pt 0pt, clip]{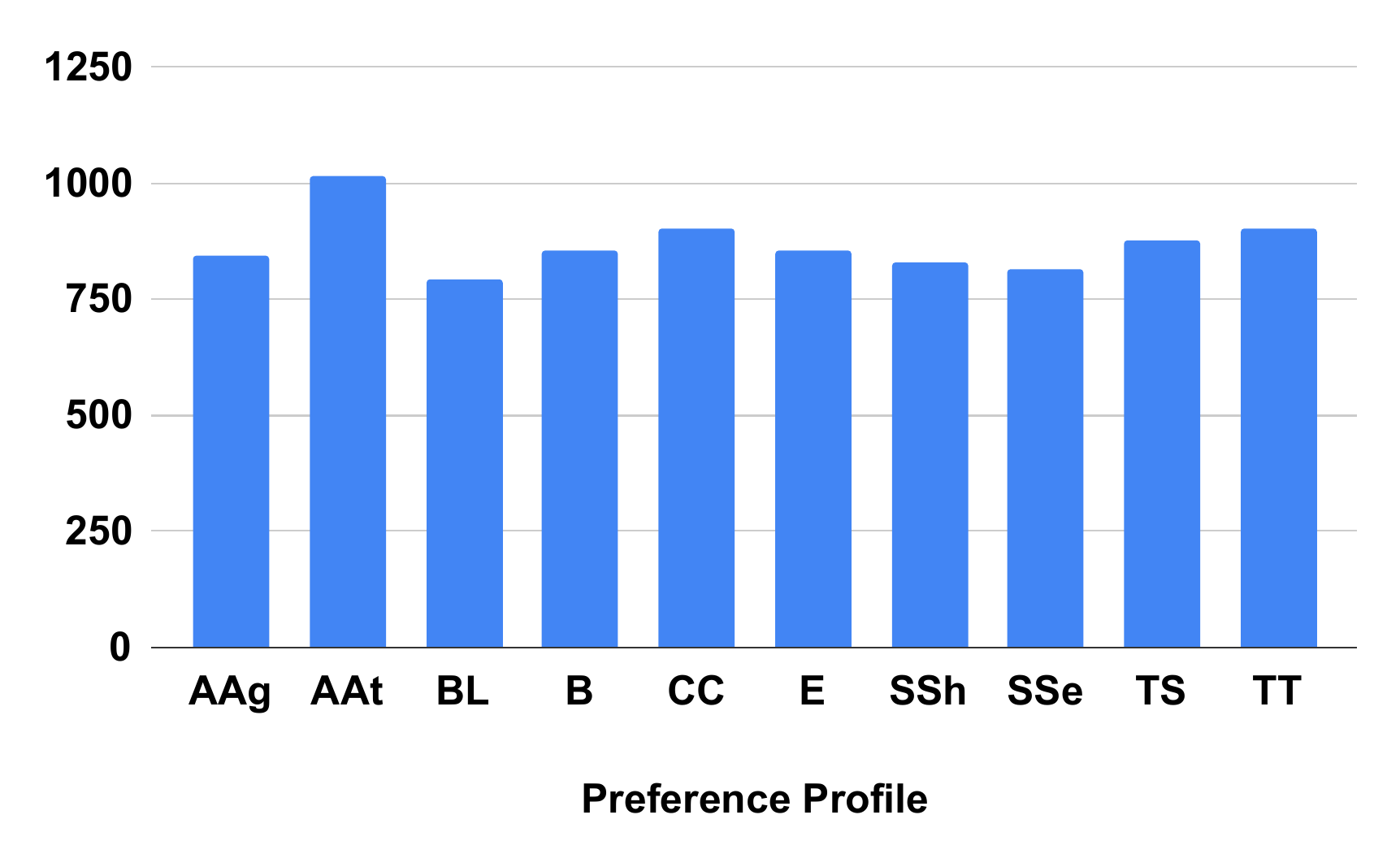}
        \caption{Preference profile distribution.}
        \label{fig:subfig1}
    \end{subfigure}
    \hspace{0.02\textwidth} 
    \begin{subfigure}[b]{0.3\textwidth}
        \centering
        \includegraphics[width=\linewidth]{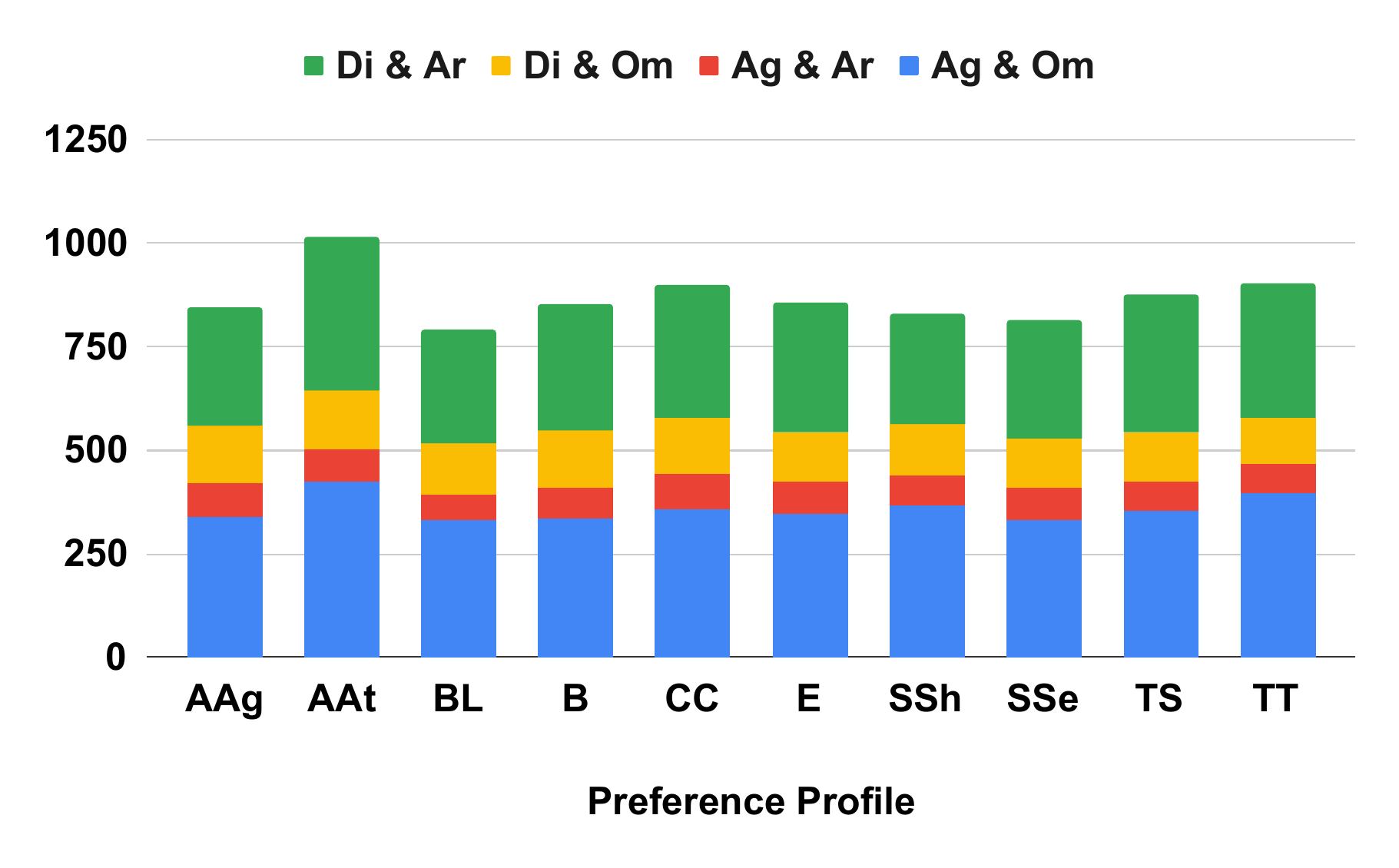}
        \caption{Argumentation profile distribution.}
        \label{fig:subfig2}
    \end{subfigure}
    \hspace{0.02\textwidth} 
    \begin{subfigure}[b]{0.3\textwidth}
        \centering
        \includegraphics[width=\linewidth]{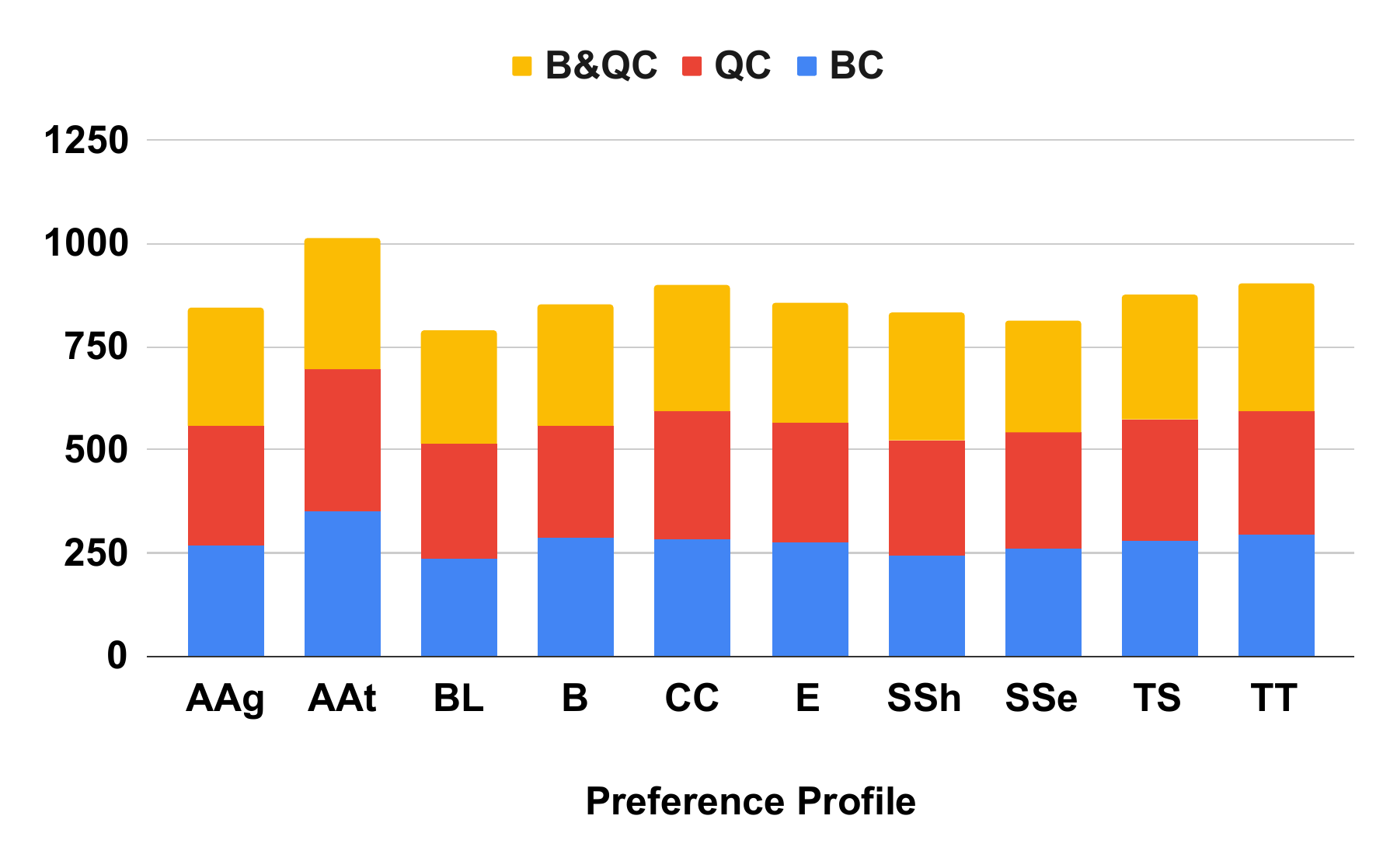}
        \caption{Buying style profile distribution.}
        \label{fig:subfig3}
    \end{subfigure}
    
    \caption{Distribution of conversations across different profile types.}
    \vspace{-12pt}
    \label{profiles_dist_main}
\end{figure*}

\noindent \textbf{Manual Assessment.} We conduct a thorough manual assessment of the filtered conversations to evaluate their quality based on the following criteria: Personality Relevance (\textbf{PR}), Bargaining Efficacy (\textbf{BE}), Outcome Fairness (\textbf{OF}), Engagingness (\textbf{E}), Coherence (\textbf{C}), Humanness (\textbf{H}), and Fluency (\textbf{F}). The assessment is conducted by four human evaluators who rate 864 randomly selected conversations on a scale of 1-4 (low to high), adhering to the specified guidelines. Appendix \ref{human_dataset} outlines assessment guidelines, criteria, and the expertise of the evaluators. The results, presented in Table \ref{manual_assessment}, indicate that the generated conversations exhibit high levels of personality relevance, effective bargaining, and fairness in negotiation outcomes, while consistently being rated as interesting, coherent, human-like, and natural, underscoring the overall high quality of the negotiation interactions. We measure the inter-evaluator agreement using Fleiss Kappa ($\kappa$) \cite{fleiss1971measuring}. The $\kappa$ scores in range of $0.4 < \kappa <  0.7$ for all criteria indicate fair agreement among evaluators \cite{mchugh2012interrater}. We provide further dataset analysis in Appendix \ref{dataset_analysis}.


\begin{table}[!hbt]\footnotesize
\centering
\begin{adjustbox}{max width=\linewidth}
\begin{tabular}{l|ccccccc}
\hline
\textbf{Criteria} & \textbf{PR} & \textbf{BE} & \textbf{OF} & \textbf{E} & \textbf{C} & \textbf{H} & \textbf{F}\\ \hline
\textbf{Value} & 3.62 & 3.48 & 3.14 & 3.75 & 3.22 & 3.07 & 3.79 \\
\textbf{$\mathbf{\kappa}$} & 0.413 & 0.458 & 0.561 & 0.698 & 0.578 & 0.687 & 0.589 \\ \hline
\end{tabular}
\end{adjustbox}

\caption{Manual assessment results of \texttt{\textbf{PACT}}.}
\label{manual_assessment}

\end{table}



\subsection{Dataset Distribution}
Table \ref{statistics_main} presents statistics for the \texttt{\textbf{PACT}} dataset. It comprises 8,687 conversations, averaging 21.43 turns per conversation with length of conversations varying from 8 to 24 turns. The dataset contains 11,522,261 tokens, averaging 61.87 tokens per turn, indicating sufficiently long conversations with extended turn lengths, contributing to complex interactions. Also, it exhibits rich lexical diversity, featuring 116,974 unique words and 545,804 unique bigrams. Figure \ref{profiles_dist_main} shows the distribution of conversations across different personality profiles. The usage frequency and proportions of each dialog act are summarized in Table \ref{acts_dist_main}. All personality profiles and dialogue acts occur in sufficient frequency, thereby ensuring a comprehensive representation of realistic conversational scenarios and demonstrating their effective utilization in the dialogue generation process. Incorporating various ABN-specific dialog acts enhances dialogue diversity and better mirrors real-life interactions. Overall, our dataset offers rich negotiation scenarios and diverse dialog acts, making it a valuable resource for advancing ABN dialogue systems. A descriptive analysis of the dataset, including (1) Diversity and Coverage Analysis, (2) Filtered Dialogue Analysis, and (3) Comparison with Existing Datasets, is given in Appendix \ref{dataset_analysis}. 


\begin{table}[!hbt]
\centering
\begin{adjustbox}{max width=\linewidth}
\begin{tabular}{l|c|l|c}
\hline
\textbf{Metrics} & \textbf{Value} & \textbf{Metrics} & \textbf{Value} \\ \hline
        \# Conversations               & 8,687   & \# Tokens            & 11,522,261 \\
        \# Turns               & 186,241 & Avg. tokens/turn           & 61.87\\
        Avg. turns/conversation                 & 21.43 & Avg. tokens/conversation          & 1,326.38\\
        Min turns       & 8 &  \# Unique words                 & 116,974\\
        Max turns          & 24 & \# Unique bigrams                  & 545,804\\
        \hline
\end{tabular}
\end{adjustbox}
\caption{Statistics of the proposed \texttt{\textbf{PACT}} dataset.}
\label{statistics_main}
\end{table}

\begin{table}[!hbt]\Huge
\centering
\begin{adjustbox}{max width=\linewidth}
\begin{tabular}{l|c|c|l|c|c}
\hline
\textbf{Dialog Acts} & \textbf{Count} & \textbf{Proportion} & \textbf{Dialog Acts} & \textbf{Count} & \textbf{Proportion} \\ \hline
        Negotiate-price-increase & 17,392 & 9.3\% & Greet-Ask & 8,687 & 4.7\%\\ 
        Negotiate-price-decrease & 36,374 & 19.5\% & Inform & 8,687 & 4.7\%\\ 
        Negotiate-price-nochange & 22,734 & 12.2\% & Elicit Preference & 8,687 & 4.7\%\\ 
        Negotiate-add-X & 14,090 & 7.6\% & Ask-clarification-X & 4,959 & 2.7\%\\ 
        Negotiate-remove-X & 20,900 & 11.2\% & Provide-clarification-X & 4,909 & 2.6\%\\ 
        Concern-price & 7,423 & 4.0\% & Ask-price & 6,759 & 3.6\%\\ 
        Disagree-price & 2,134 & 1.1\% & Tell-price & 6,578 & 3.5\%\\ 
        Justify-price & 2,149 & 1.2\% & Provide-consent & 3,624 & 1.9\%\\ 
        Assurance-price & 2,153 & 1.2\% & Consent-response & 3,476 & 1.9\%\\ 
        Disagree-X & 1,014 & 0.5\% & Accept & 2,507 & 1.3\%\\ 
        Justify-X & 197 & 0.1\% & Acknowledge-acceptance & 553 & 0.3\%\\ 
        Assurance-X & 255 & 0.1\% & & & \\\hline
\end{tabular}
\end{adjustbox}
\caption{ABN-specific dialog acts distribution in \texttt{\textbf{PACT}}.}
\label{acts_dist_main}
\end{table}

\begin{table*}[!hbt]\Large
\centering
\begin{adjustbox}{max width=0.95\linewidth}
\begin{tabular}{c|l|cc|cc|ccccccccccccc} \hline
\multirow{2}{*}{\textbf{Task Setup}} & \multirow{2}{*}{\textbf{Models}} & \multicolumn{2}{c|}{\textbf{PR}}                        & \multicolumn{2}{c|}{\textbf{DAP}}                       & \multicolumn{11}{c}{\textbf{RG}}                                                                                                                                                                                                                                                                                                                      \\ \cline{3-19}
 &  & \textbf{Acc $\uparrow$}         & \multicolumn{1}{c|}{\textbf{F1 $\uparrow$}} & \textbf{Acc $\uparrow$}         & \multicolumn{1}{c|}{\textbf{F1 $\uparrow$}} & \textbf{PPL $\downarrow$}         & \multicolumn{1}{c}{\textbf{B-1 $\uparrow$}} & \multicolumn{1}{c}{\textbf{B-2 $\uparrow$}} & \multicolumn{1}{c}{\textbf{MET $\uparrow$}} & \multicolumn{1}{c}{\textbf{R-L $\uparrow$}} & \multicolumn{1}{c}{\textbf{D-1 $\uparrow$}} & \multicolumn{1}{c}{\textbf{D-2 $\uparrow$}} & \multicolumn{1}{c}{\textbf{BS-F1 (\%) $\uparrow$}} & \multicolumn{1}{c}{\textbf{R-LEN $\uparrow$}} & \multicolumn{1}{c}{\textbf{P-ACC $\uparrow$}} & \multicolumn{1}{c}{\textbf{D-ACC $\uparrow$}} & \multicolumn{1}{c}{\textbf{FEQA $\uparrow$}} & \multicolumn{1}{c}{\textbf{QE $\uparrow$}} \\\hline
\multirow{6}{*}{PMs} 
        & Llama-2 & 0.269 & 0.296 & 0.302 & 0.307 & 69.23 & 0.133 & 0.076 & 0.229 & 0.194 & 0.041 & 0.254 & 63.24 & 25.92 & 0.288 & 0.459 & 18.98 & 21.24 \\
        & Llama-2-chat & 0.278 & 0.300 & 0.320 & 0.299 & 65.99 & 0.141 & 0.083 & 0.245 & 0.204 & 0.045 & 0.265 & 64.78 & 27.68 & 0.302 & 0.472 & 19.64 & 22.71 \\
        & Llama-3 & 0.285 & 0.310 & 0.325 & 0.319 & 66.73 & 0.150 & 0.088 & 0.252 & 0.210 & 0.049 & 0.278 & 65.74 & 29.16 & 0.312 & 0.479 & 21.02 & 24.12 \\
        & Llama-3.1 & 0.291 & 0.316 & 0.336 & 0.326 & 64.42 & 0.159 & 0.096 & 0.270 & 0.221 & 0.052 & 0.287 & 66.92 & 31.31 & 0.321 & 0.493 & 22.38 & 25.29 \\
        & Mistral & 0.266 & 0.287 & 0.307 & 0.298 & 68.16 & 0.130 & 0.080 & 0.241 & 0.186 & 0.043 & 0.257 & 64.18 & 26.92 & 0.293 & 0.469 & 19.85 & 22.04 \\
        & Vicuna & 0.288 & 0.304 & 0.316 & 0.304 & 64.46 & 0.143 & 0.093 & 0.250 & 0.198 & 0.048 & 0.270 & 65.49 & 29.67 & 0.305 & 0.481 & 20.67 & 23.71 \\
        \hline
        \multirow{6}{*}{PMs\textit{w}P} 
        & Llama-2 & - & - & - & - & 54.30 & 0.174 & 0.122 & 0.293 & 0.238 & 0.061 & 0.319 & 67.82 & 34.32 & 0.361 & 0.513 & 23.19 & 25.87 \\
        & Llama-2-chat & - & - & - & - & 52.12 & 0.189 & 0.130 & 0.308 & 0.247 & 0.065 & 0.325 & 68.42 & 35.84 & 0.374 & 0.526 & 24.48 & 27.13 \\
        & Llama-3 & - & - & - & - & 50.71 & 0.195 & 0.139 & 0.314 & 0.256 & 0.067 & 0.338 & 69.58 & 37.53 & 0.382 & 0.533 & 25.25 & 28.36 \\
        & Llama-3.1 & - & - & - & - & 49.12 & 0.201 & 0.148 & 0.323 & 0.265 & 0.070 & 0.346 & 70.74 & 39.24 & 0.390 & 0.542 & 26.31 & 29.52 \\
        & Mistral & - & - & - & - & 52.89 & 0.184 & 0.123 & 0.304 & 0.238 & 0.059 & 0.322 & 67.31 & 34.97 & 0.365 & 0.517 & 23.87 & 26.45 \\
        & Vicuna & - & - & - & - & 51.24 & 0.192 & 0.132 & 0.315 & 0.246 & 0.065 & 0.333 & 68.92 & 36.62 & 0.370 & 0.525 & 24.54 & 28.01 \\
        \hline
        \multirow{6}{*}{FT-\textit{H2HNego}} & Llama-2 & 0.385 & 0.402 & 0.422 & 0.415 & 45.62 & 0.224 & 0.157 & 0.338 & 0.274 & 0.078 & 0.373 & 79.48 & 41.52 & 0.412 & 0.560 & 33.27 & 36.42 \\
        & Llama-2-chat & 0.397 & 0.413 & 0.436 & 0.423 & 43.72 & 0.231 & 0.165 & 0.346 & 0.287 & 0.082 & 0.379 & 80.11 & 43.02 & 0.424 & 0.573 & 34.61 & 37.71 \\
        & Llama-3 & 0.409 & 0.425 & 0.443 & 0.434 & 42.51 & 0.239 & 0.172 & 0.358 & 0.291 & 0.086 & 0.385 & 81.29 & 44.91 & 0.435 & 0.585 & 36.02 & 38.92 \\
        & Llama-3.1 & 0.417 & 0.432 & 0.453 & 0.448 & 41.32 & 0.247 & 0.182 & 0.368 & 0.303 & 0.090 & 0.393 & 82.41 & 46.68 & 0.446 & 0.592 & 37.16 & 40.08 \\
        & Mistral & 0.390 & 0.407 & 0.419 & 0.415 & 45.89 & 0.223 & 0.152 & 0.334 & 0.269 & 0.075 & 0.369 & 79.92 & 42.11 & 0.414 & 0.566 & 33.98 & 36.46 \\
        & Vicuna & 0.402 & 0.420 & 0.437 & 0.425 & 41.75 & 0.234 & 0.166 & 0.344 & 0.285 & 0.082 & 0.382 & 81.03 & 44.24 & 0.429 & 0.578 & 35.13 & 37.83 \\
        \hline
        \multirow{6}{*}{FT-\textit{Vanilla}} & Llama-2 & 0.455 & 0.472 & 0.473 & 0.465 & 51.65 & 0.29  & 0.183 & 0.406 & 0.306 & 0.074 & 0.426 & 83.23 & 49.66 & 0.507 & 0.651 & 44.23 & 46.72 \\
        & Llama-2-chat & 0.460 & 0.466 & 0.481	& 0.478 & 39.62 & 0.299 & 0.186 & 0.423 & 0.308 & 0.083 & 0.418 & 83.83 & 57.14 & 0.535 & 0.677 & 45.98 & 47.85 \\
        & Llama-3 & 0.475 & 0.464 & 0.473	& 0.466 & 39.37 & 0.272 & 0.161 & 0.39  & 0.288 & 0.073 & 0.395 & 83.19 & 48.84 & 0.478 & 0.667 & 43.12 & 45.01 \\
        & Llama-3.1 & 0.450 & 0.432 & 0.505	& 0.498 & 35.59 & 0.292 & 0.179 & 0.414 & 0.302 & 0.082 & 0.425 & 83.68 & 57.82 & 0.534 & 0.675 & 46.34 & 48.39 \\
        & Mistral & 0.460 & 0.459 & 0.468 &	0.487 & 46.64 & 0.168 & 0.092 & 0.316 & 0.181 & 0.074 & 0.429 & 77.54 & 29.21 & 0.494 & 0.652 & 42.18 & 43.79 \\
        & Vicuna & 0.470 & 0.478 & 0.491 &	0.479 & 36.25 & 0.319 & 0.211 & 0.437 & 0.337 & 0.086 & 0.442 & 84.56 & 54.18 & 0.574 & 0.693 & 47.56 & 49.81 \\
        \hline
        \multirow{6}{*}{FT-\textit{Ours}} & Llama-2 & 0.512 & 0.528 & 0.567 & 0.548 & 18.65 & 0.375 & 0.245 & 0.512 & 0.401 & 0.098 & 0.525 & 87.35 & 59.24 & 0.723 & 0.791 & 49.82 & 52.69 \\
        & Llama-2-chat & 0.530 & 0.546 & 0.583 & 0.563 & 14.25 & 0.358 & 0.229 & 0.508 & 0.393 & 0.105 & 0.523 & 88.23 & 62.01 & 0.728 & 0.813 & 51.16 & 53.94 \\
        & Llama-3 & 0.524 & 0.539 & 0.578 & 0.558 & 15.92 & 0.361 & 0.238 & 0.495 & 0.387 & 0.099 & 0.514 & 88.14 & 61.48 & 0.703 & 0.767 & 49.35 & 51.89 \\
        & Llama-3.1 & 0.518 & 0.533 & 0.572 & 0.553 & 17.82 & 0.389 & 0.258 & 0.527 & 0.419 & 0.102 & 0.538 & 87.92 & 60.11 & 0.739 & 0.802 & 50.43 & 53.27 \\
        & Mistral & 0.515 & 0.525 & 0.559 & 0.538 & 16.87 & 0.279 & 0.157 & 0.392 & 0.289 & 0.102 & 0.472 & 82.67 & 35.44 & 0.692 & 0.782 & 47.91 & 50.23 \\
        & Vicuna & \textbf{0.537} & \textbf{0.550} & \textbf{0.592} & \textbf{0.571} & \textbf{13.98} & \textbf{0.412} & \textbf{0.275} & \textbf{0.552} & \textbf{0.431} & \textbf{0.111} & \textbf{0.562} & \textbf{89.23} & \textbf{63.27} & \textbf{0.789} & \textbf{0.841} & \textbf{52.74} & \textbf{55.13} \\
        \hline
        & GPT-4.1-mini & 0.425 & 0.438 & 0.478 & 0.459 & 16.85 & 0.385 & 0.248 & 0.515 & 0.397 & 0.104 & 0.457 & 78.26 & 41.92 & 0.725 & 0.746 & 39.38 & 34.02 \\ \hline
\end{tabular}
\end{adjustbox}
\caption{\footnotesize Automatic evaluation results. All results are statistically significant based on Welch’s t-test \cite{welch1947generalization} conducted at a 5\% significance level. Best scores are in bold.}
\label{auto_res_rg}
\end{table*}

\vspace*{-1em}
\section{Experiments}

\textbf{Tasks.} 
Upon the construction of \texttt{\textbf{PACT}}, we can support an entire PAN-DG task mirroring the real-life negotiation scenario. We split the entire personality-driven ABN dialogue generation process into 3 sub-tasks: Personality Recognition (PR) that aims to identify the personality profiles of interlocutors through dialogue context; Dialog Act Prediction (DAP) that aims to infer the future dialog act for a given dialogue context; and Response Generation (RG) that aims to generate travel agents’ probable response based on the dialogue context. To ensure factual consistency and mitigate hallucinations for RG task, we integrate background knowledge through a Retrieval-Augmented Generation (RAG) technique \cite{lewis2020retrieval}. Specifically, we leverage the curated package database containing verified travel details, including a comprehensive list of amenities, services, and their corresponding pricing (\textsection \ref{PACT_generation}). 

Further, to align responses with personality and dialogue act, we integrate outputs from the PR and DAP sub-tasks. We use a multi-task learning (MTL) approach during training, where the model learns the relationships between personality, dialogue act, and response generation. During inference, PR and DAP provide dynamic personality and dialog act information for contextually accurate, personality-driven response generation. We discuss RAG and MTL in detail in Appendix \ref{response_generation}.  

\noindent \textbf{Models.} We evaluate 7 LLMs - LLaMA models (Llama-2-7b, Llama-2-7b-chat-hf, Llama-3-8b, Llama-3.1-8b) \cite{touvron2023llama}, along with Mistral-8b \cite{jiang2023mistral}, Vicuna-7b \cite{chiang2023vicuna}, and GPT-4.1-mini \cite{openai2024chatgpt}, across three sub-tasks. 


\noindent \textbf{Task Setups. }Following \cite{han2024psydial}, we evaluate the effectiveness of \texttt{\textbf{PACT}} on the PAN-DG task under five setups: (i) Pre-trained Models (PMs), (ii) Pre-trained Models with Personality (PMs\textit{w}P), (iii) Fine-tuning on a Human-Human Negotiation Dialogue Dataset (FT-\textit{H2HNego}), (iv) Fine-tuning on \texttt{\textbf{PACT}} (FT-\textit{Vanilla}), and (v) Fine-tuning on \texttt{\textbf{PACT}} with background knowledge, personality, and dialog act information (FT-\textit{Ours}). We follow a 70:20:10 train-validation-test split for all setups. Setups details are given in Appendix \ref{task_setup}.

\noindent \textbf{Evaluation Metrics.} For PR and DAP tasks, we use Accuracy (Acc) and Macro F1 (F1). For RG task, we employ Perplexity (PPL), BLEU-1/2 (B-1/2), METEOR (MET), ROUGE-L (R-L), DISTINCT-1/2 (D-1/2), BERTScore-F1 (BS-F1), Response Length (R-LEN), Personality Accuracy (P-ACC), and Dialog Act Accuracy (D-ACC). Further, to measure faithfulness of generated response with dialogue context and package information, we employ FEQA and QuestEval (QE). Details on task setups, implementation specifications, and evaluation metrics are given in Appendix \ref{experiments}.

\section{Evaluation Results}

Table \ref{auto_res_rg} reports LLM performance across tasks and setups. Pre-trained models perform poorly on PR, DAP, and RG tasks, lacking sensitivity to personality and negotiation nuances. Fine-tuning on \texttt{\textbf{PACT}} yields notable gains, matching or outperforming fine-tuning on human-human negotiation data, for instance, Vicuna achieves 84.56\% BS-F1 on \texttt{\textbf{PACT}} vs. 81.03\% on FT-\textit{H2HNego}, establishing \texttt{\textbf{PACT}} as a strong fine-tuning alternative.

Our proposed approach (FT-\textit{Ours}), which incorporates package knowledge along with personality and dialog act information, shows the highest improvements across all sub-tasks. Notably, for RG task, it shows substantial improvements across multiple metrics, including response quality, dialogue coherence, and factual accuracy. A significant reduction in perplexity reflects enhanced model confidence and fluency. Moreover, integrating personality and dialog act information results in increased P-ACC and D-ACC. Specifically, compared to FT-\textit{Vanilla} counterpart, it obtains a 27.2\% increase in P-ACC and an 17.6\% increase in D-ACC for Vicuna, highlighting the importance of these aspects. The notable improvements in FEQA and QE scores further underscore enhanced factual grounding and response relevance. These results demonstrate that background package knowledge reduces the generation of generic responses while improving relevance and accuracy. 

Furthermore, compared to ChatGPT, Vicuna under our proposed setup (FT-Ours) achieves significantly better performance across all tasks, with a 20.4\% increase in F1 for PR, 19.6\% for DAP, and 12.3\% in BS-F1 for RG. Despite ChatGPT’s strong general language capabilities, it underperforms on the PAN-DG task, highlighting the necessity of domain-specific fine-tuning. The low scores on PR and DAP suggest limited sensitivity to personality traits and dialog acts, which are critical for structured negotiation. Its weaker RG performance further reflects challenges in generating contextually relevant, factual, dialog-act, and persona-driven responses. Overall, these findings show that our dataset significantly improves the model’s ability to generate responses that not only reflect personality but also align with structured dialog dynamics, thereby emphasizing the importance of structured dialog acts, personality adaptation, and knowledge enrichment in developing advanced, personality-driven ABN dialogue systems. Human evaluations for RG are provided in Appendix \ref{human_eval}.

\section{Conclusion}


This work introduces a novel Personality-driven Argumentation-based Negotiation Dialogue Generation (PAN-DG) task, focused on generating personality-driven negotiation dialogues in the tourism domain. For PAN-DG task, we curate \texttt{\textbf{PACT}} dataset through a multi-stage pipeline using LLMs. \texttt{\textbf{PACT}} integrates multi-dimensional personality profiles and ABN-specific dialog acts to enhance personalized negotiation interactions. We also establish strong LLM baselines for this task.


\section*{Limitations}
\textbf{Controllability.} In this study, we outline a prompting technique for Large Language Models (LLMs) to generate the \texttt{\textbf{PACT}} dataset, a process that requires substantial computational resources and advanced LLMs. Our approach demonstrates promising results in terms of several manual assessment criteria. However, the quality of the generated dataset is inherently limited by the capabilities of the LLM, as prompting remains an uncontrolled generation process. Future research could explore incorporating additional controls in the prompt or employing weak supervision techniques, as demonstrated by \citet{chen2022weakly}, to further improve the quality of synthetic data.

\noindent \textbf{Negotiation Setting.} A notable limitation of our method lies in its constrained ability to fully replicate realistic negotiation conversations, as it assumes participants are fluent, the conversation flow is coherent, and there are no interruptions caused by external factors or technical issues. Moreover, the approach faces challenges in capturing subtle cultural nuances in personality profiles, such as humor or user-specific traits, which often become apparent only through repeated interactions. Besides, the scope of our work is restricted to the tourism negotiation domain, which involves complex, multi-faceted negotiations, such as balancing customer preferences, needs, pricing, and services, thereby providing a rich context for evaluating the effectiveness of ABN in fostering collaborative and mutually beneficial outcomes. 

Further, while our current framework models negotiations as self-contained sessions, consistent with many real-world tourism scenarios, we acknowledge that real-world negotiations are often iterative, occurring over multiple interactions. Extending our framework to support such multi-session dialogues represents a promising future direction. Also, our current approach models traveler personality using a single dominant trait, which may oversimplify real-world behavior. Future work could explore incorporating multi-faceted personalities that reflect combinations of traits, enabling more nuanced preference modeling and improving the system's effectiveness in complex negotiation scenarios. Nevertheless, the proposed method holds significant potential for adaptation to other negotiation domains, offering opportunities for broader applicability and deeper investigation into diverse negotiation scenarios.

\noindent \textbf{Prompt Design.} In this work, we do not claim that our prompt design is novel, nor do we assert that it is the optimal method for generating negotiation conversations. Our prompt is structured in a conversational way, inspired by the work of \citet{chen2022weakly}. Rather than focusing on the novelty of our prompt, we emphasize its application in synthesizing argumentation-based negotiation dialogue datasets. The idea of generating argumentation-based negotiation dialogues from scratch remains underexplored and presents significant opportunities to advance research in argumentation-based negotiation dialogue systems.

\noindent \textbf{Computational Costs and Reproducibility.} It is important to note that we used the publicly accessible free-tier API of LLM \texttt{Gemini-1.5-Flash} for generating negotiation dialogues due to the financial and computational limitations commonly faced in academic settings. While LLMs like ChatGPT and other advanced models have achieved significant progress in generating fluent and coherent responses, they are not a one-size-fits-all solution. These models also struggle with personalization, primarily due to the lack of diverse datasets and limitations in effectively modeling user personas \cite{chen2024recent,han2024psydial}. Nevertheless, research has demonstrated that LLMs are capable of embodying personas or personality traits when provided with appropriately designed prompts \cite{jiang2023personallm,salewski2024context}. In this context, achieving satisfactory performance using smaller (fewer parameters) or openly accessible LLMs presents several advantages: (i) these models are substantially more cost-effective, enabling broader participation in research and development even within resource-constrained settings; (ii) their open accessibility ensures reproducibility, a cornerstone of academic research, as such models can be freely shared and reimplemented by others to validate findings and extend the work further. This approach not only addresses resource constraints but also aligns with the academic objective of fostering inclusive and collaborative progress in the field. Thus, simply relying on closed-source models like GPT-4 \cite{openai2024chatgpt} or other closed-source powerful LLMs for dataset generation, response generation or personality recognition is neither practical nor essential when smaller, openly available LLMs can deliver comparable outcomes in a cost-efficient and reproducible manner. Nevertheless, future studies could examine the effectiveness of using more advanced, closed-source LLMs, such as GPT-4 \cite{openai2024chatgpt} or Gemini-1.5-Pro \cite{team2023gemini}, to enhance the generation of negotiation dialogue datasets.

\section*{Ethics Statement}
\textbf{Datasets.} Ethical considerations play a critical role in the construction and use of the dataset. Given that the dialogues are synthetically generated, it is important for users to engage with the data responsibly, with awareness of the potential risks, such as cultural appropriation or misrepresentation. Since language models are trained on web-based data, there is a risk of perpetuating biases that may reinforce stereotypes, contribute to discrimination, or marginalize certain communities. Previous research has shown that synthetic data can exacerbate these biases, leading to feedback loops that increase the likelihood of generating biased language \cite{taori2023data}. Hence, we collaborate with linguists, and language experts, who are core members of our research team to filter and manually evaluate generated conversation to prevent unintentional perpetuation of stereotypes or cultural insensitivity. Human evaluators are compensated in accordance with institutional policies. To safeguard proper usage, access to the dataset will be granted only upon signing an agreement that stipulates its use solely for research purposes. 

\textbf{Negotiation Systems.} In light of the inherent bargaining dynamics present in negotiations, the development and deployment of argumentation-based negotiation dialogue systems must be guided by strict ethical standards. This study was reviewed and approved by our Institutional Review Board (IRB). Our approach to personality-driven argumentation-based negotiation emphasizes user flexibility and aims to foster win-win, mutually beneficial outcomes. Unlike zero-sum negotiations, where one party’s gain is another’s loss, our system is designed to maximize benefits for both parties. Crucially, users retain full autonomy, ensuring they are never pressured into accepting deals that do not meet their needs.

\textbf{LLMs.} We utilized large language models (LLMs) exclusively within their designated frameworks, adhering to all relevant licensing agreements. Our approach aligns with a range of contemporary initiatives focused on assessing the varied capabilities of LLMs in generating synthetic conversations, thereby guaranteeing compliance with ethical standards and operational protocols.

\section*{Acknowledgement}
Priyanshu Priya acknowledges the financial support provided by the Department of Science and Technology, Ministry of Science and Technology, Government of India, through the Innovation in Science Pursuit for Inspired Research (INSPIRE) Fellowship. The research reported in this paper is an outcome of the project titled \enquote{Conversational Agents with Negotiation and Influencing Ability}, sponsored by Accenture Labs, Bangalore, India.

\bibliography{custom,manuscript}

\newpage
\appendix

\section*{Appendix}

\section{Travel Packages}
\label{package}

The database consists of information about 10 distinct travel packages, namely Tranquil Escape, Sight Tour, Collector's Haven, Adrenaline Rush, Vibrant Nightlife, Cultural Odyssey, Nautical Adventure, Active Pursuits, Nature Wanderer, Sandy Serenity. A sample travel package information is given in Table \ref{package_info}.

\begin{table*}[!hbt]
\centering
\begin{adjustbox}{max width=\linewidth}
\begin{tabular}{l|p{15cm}}
\hline
\textbf{Metrics} & \textbf{Description} \\ \hline
Personality Relevance & Examines whether generated conversation is consistent with the given traveler's and travel agent's personality.\\
Bargaining Efficacy & Measures how effectively each party present offers, compelling arguments, reasoning, or concessions that influence the opponent’s decision-making process. \\
Outcome Fairness & Examines the equity of the final negotiation outcome, determining how effectively it achieves a win-win scenario.\\
Engagingness &  Assesses the ability of the dialogues to sustain the user’s interest throughout the interaction.\\
Coherent & Evaluates how logically structured and clear the overall conversation is, with particular attention to the dialogue flow and connection between the utterances.\\
Humanness & Evaluates how closely the generated conversation resembles a human conversation. \\
Naturalness & Evaluates how closely the generated conversation resembles a natural conversation. \\ \hline

\end{tabular}
\end{adjustbox}
\caption{Description of manual assessment criteria.}
\label{manual_metrics}
\end{table*}

\begin{table*}[!hbt]\footnotesize
\begin{center}
\begin{adjustbox}{max width=\linewidth}
\begin{tabular}{lcc}
\hline
\textbf{Budget}        & \textbf{Price Range}                                              & $c$ \\ \hline
Low Budget             & traveler initial price <= 0.65 $\times$ agent initial price        & 1.2        \\
High Budget            & traveler initial price >= 0.85 $\times$ agent initial price        & 0.6        \\
Moderate Budget        & 0.85 $\times$ agent initial price > traveler initial price > 0.65 $\times$ agent initial price & 0.9        \\ \hline
\end{tabular}
\end{adjustbox}
\end{center}
\caption{Based on the initial prices proposed by the traveler and the agent, the initial price gap is determined. A significant price gap is classified as a low-budget scenario, while a smaller gap corresponds to a high-budget scenario; intermediate gaps are categorized as moderate-budget scenarios. The accompanying table presents the relationship between the traveler's initial price and the agent's initial price across these three budget scenarios. Additionally, the table includes the corresponding values of the concession factor ($c$) for each budget category: low, moderate, and high.}
\label{value_c}
\end{table*}

\begin{figure*}[hbt!]
\centering
    \includegraphics[width = \linewidth,scale=1.0]{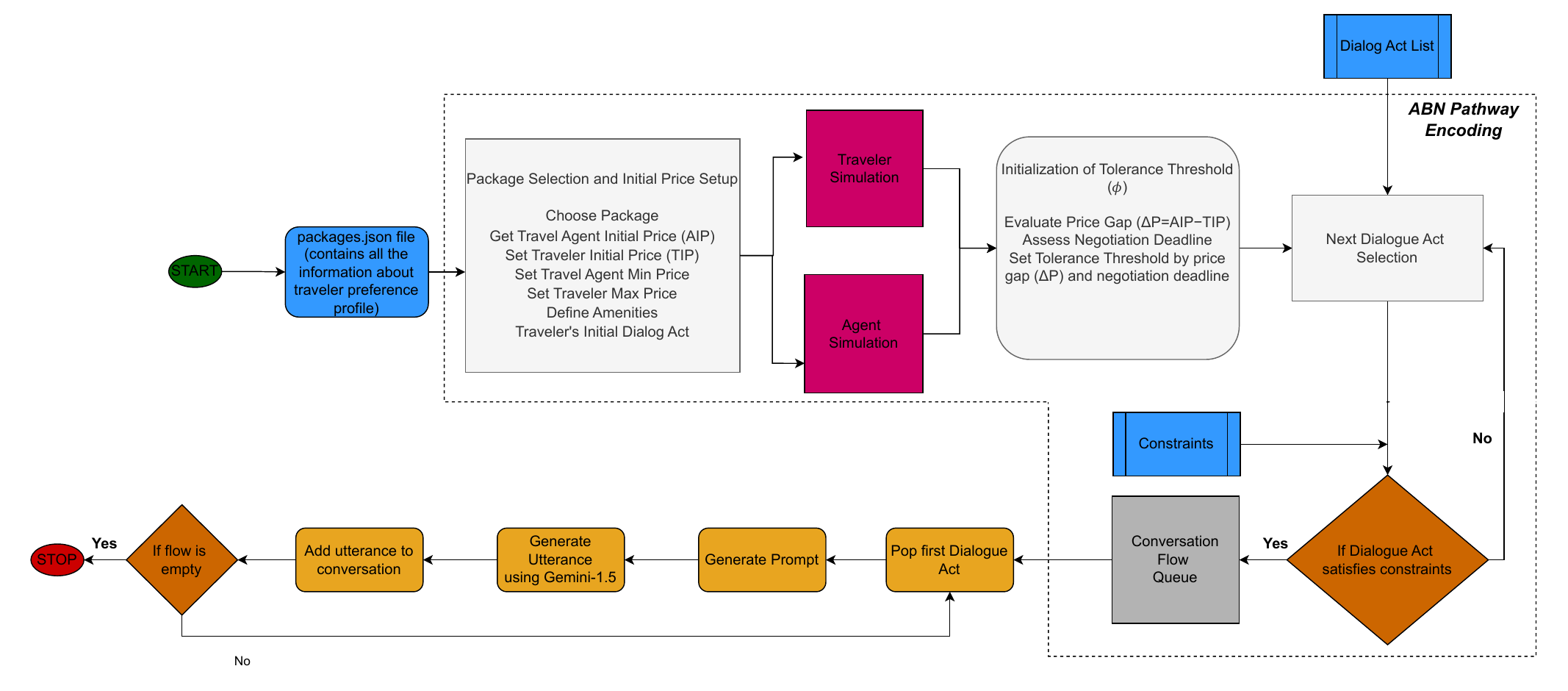}
    \caption{Flow chart illustrating \texttt{\textbf{PACT}} dataset generation process.} 
    \label{dataset_creation}
\end{figure*}

\section{Personality Profiles Description}
\label{profile_desc}

\subsection{Argumentation Profiles}

\begin{itemize}[leftmargin=*,topsep=0pt]
\setlength{\itemsep}{0pt}
\setlength{\parskip}{0pt}
\setlength{\parsep}{0pt}
    \item \textbf{Agreeable} characterizes interlocutors who tend to accept offers and arguments with minimal conflict, promoting a cooperative and consensus-driven approach. They prioritize reaching mutual agreements and are less likely to challenge or reject proposals, demonstrating flexibility and accommodation throughout the negotiation process.

    \item \textbf{Disagreeable} characterizes interlocutors adopting a more critical and skeptical stance toward offers and arguments. They tend to reject proposals unless presented with compelling reasons to accept them. They challenge positions frequently, leading to more difficult negotiations as they prioritize ensuring their own interests are satisfied.

    \item \textbf{Open-minded} characterizes interlocutors engaging constructively in negotiation, critically evaluating offers and arguments without immediate acceptance or rejection. They strike a balance between cooperation and skepticism, accepting or countering arguments based on rational evaluation. These agents focus on maintaining a productive dialogue and are more willing to explore alternative solutions.

    \item \textbf{Argumentative} characterizes interlocutors who indicate a strong tendency to challenge and counter opposing offers or arguments. They are more confrontational and are quick to reject or attack propositions, actively engaging in debates to assert their own position. Their negotiation pattern often involves intense argumentation, seeking to dominate the dialogue through persuasive counterarguments.
\end{itemize}

\subsection{Preference Profiles}
\begin{itemize}[leftmargin=*,topsep=0pt]
\setlength{\itemsep}{0pt}
\setlength{\parskip}{0pt}
\setlength{\parsep}{0pt}
\item \textbf{Culture Creature} prefers cultural experiences during trips, drawn to theater performances, museums, monuments, art exhibitions, and local festivals. 
\item \textbf{Action Agent} prefers dynamic and lively environments, seeking out nightclubs, upscale restaurants, and various entertainment venues to enhance their travel experience.
\item \textbf{Avid Athlete} prefers staying active during vacations, focusing on engaging in favorite sports such as golf, tennis, or other athletic activities. 
\item \textbf{Thrill Seeker} prefers adrenaline-pumping, high-risk activities that provide excitement and an emotional high, such as skydiving, bungee jumping, or extreme sports. 
\item \textbf{Trail Trekker} prefers outdoor experiences, actively seeking opportunities for hiking, walking, exploring parks and forests, and enjoying nature in mountains or through birdwatching. 
\item \textbf{Escapist} prefers vacations that offer a retreat from everyday life, seeking tranquility and relaxation in quiet, peaceful environments. 
\item \textbf{Shopping Shark} prefers destinations with vibrant shopping opportunities, including shopping areas and local markets. 
\item \textbf{Boater} prefers water-based destinations where their boat serves as a home, valuing the experience of being on the water and exploring coastal or lakefront areas.
\item \textbf{Sight Seeker} prefers to explore and pause for landmarks, events, scenic vistas, and attractions, always eager to take in the sights and experiences along the journey. 
\item \textbf{Beach Lover} prefers to relax and sunbathe in warm destinations characterized by ample sun, soft sand, and inviting water.
\end{itemize}

\subsection{Buying Style Profiles}

\begin{itemize}[leftmargin=*,topsep=0pt]
\setlength{\itemsep}{0pt}
\setlength{\parskip}{0pt}
\setlength{\parsep}{0pt}
\item \textbf{Quality-concerned traveler} prioritizes high standards in amenities and services, seeking assurance of superior quality regardless of cost.
\item \textbf{Budget-concerned traveler} seeks cost-effective options, emphasizing value for money and actively comparing prices to maximize their budget.
\item \textbf{Budget-\&-Quality-concerned traveler} balances quality and cost, seeking well-reviewed options that meet both quality standards and budget constraints.
\end{itemize}

\section{ABN-specific Dialog Acts Definition}
\label{acts_desc}

\subsection{Negotiation Dialog Acts}
\begin{itemize}[leftmargin=*,topsep=0pt]
\setlength{\itemsep}{0pt}
\setlength{\parskip}{0pt}
\setlength{\parsep}{0pt}
    \item \textbf{Negotiate-price-increase:} Used by the travel agent to negotiate a higher price for a package or its related features and/or services during the deal.
    \item \textbf{Negotiate-price-decrease:} Used by the traveler to negotiate a lower price for a package or service.
    \item \textbf{Negotiate-price-nochange:} Suggests that the price of a package or service remains unchanged, emphasizing its current value.
    \item \textbf{Negotiate-add-X:} Proposes the addition of a particular feature and/or service during the negotiation. 
    \item \textbf{Negotiate-remove-X:} Suggests the removal of a specific feature and/or service from the deal, potentially affecting the price.
\end{itemize}

\subsection{Argumentation Dialog Acts}
\begin{itemize}[leftmargin=*,topsep=0pt]
\setlength{\itemsep}{0pt}
\setlength{\parskip}{0pt}
\setlength{\parsep}{0pt}
    \item \textbf{Concern-price:} Expresses hesitation regarding the proposed price, possibly indicating dissatisfaction or a need for further discussion.
    \item \textbf{Disagree-price:} Expresses contention to the proposed price, often leading to further negotiation.
    \item \textbf{Justify-price:} Provides reasoning to support the proposed price to reinforce its fairness or value.
    \item \textbf{Assurance-price:} Reassures that the price is reasonable, thereby addressing any concerns or doubts raised during the negotiation. 
    \item \textbf{Disagree-X:} Raises objections to a specific feature or term, indicating that changes are necessary for agreement.
    \item \textbf{Justify-X:} Defends the inclusion of a specific feature by presenting logical reasoning or benefits to support its value. 

    \item \textbf{Assurance-X:} Offers reassurance about the validity or quality of a specific feature to mitigate concerns or doubts about its inclusion. 
\end{itemize}

\subsection{General Dialog Acts}
\begin{itemize}[leftmargin=*,topsep=0pt]
\setlength{\itemsep}{0pt}
\setlength{\parskip}{0pt}
\setlength{\parsep}{0pt}
    \item \textbf{Greet-Ask:} Opening the conversation politely and simultaneously requesting specific details about a feature, service, travel package, or current negotiation. 
    \item \textbf{Inform:} Sharing comprehensive information related to the packages or services under negotiation.
    \item \textbf{Elicit-preference:} Expressing the traveler's individual preferences regarding package features, services, quality, and/or budget. 
    \item \textbf{Ask-price:} Seeking clarification on  pricing of a package or its related features and/or services.
    \item \textbf{Tell-price: }Stating the proposed price of a package or its related features and/or services.
    \item \textbf{Ask-clarification-X: } Requesting further clarification on particular aspect of the negotiation or the entire deal in line with the current negotiation context.
    \item \textbf{Provide-clarification-X: }Offering clarifications about certain aspects of the negotiation or the entire deal.
    \item \textbf{Provide-consent:} Expressing agreement or approval of a proposal   
    \item \textbf{Consent-response:} Responding to the consent given, indicating that the agent understands the traveler’s approval and is ready to proceed.
    \item \textbf{Accept:} Indicating agreement to the offer, proposal, or deal presented.
    \item \textbf{Acknowledge-acceptance: }Recognizing the acceptance of the deal by the other party.
\end{itemize}

\section{Preliminary Analysis for LLM Selection}
\label{llm_selection}
To select the LLM for our work, we initially generated 50 conversations using \texttt{LLaMA-2-7B}, \texttt{Mistral-7B}, and \texttt{Gemini-1.5-Flash}, and then manually assessed them for \textit{Personality Relevance (PR)}, \textit{Bargaining Efficacy (BE)}, \textit{Outcome Fairness (OF)}, \textit{Engagingness (E)}, \textit{Coherence (C)}, \textit{Humanness (H)}, and \textit{Fluency (F)}. These dialogues are evaluated by four human annotators using a 4-point Likert scale. The average inter-evaluator agreement scores were 0.678 for \texttt{LLaMA-2-7B}, 0.672 for \texttt{Mistral-7B}, and 0.711 for \texttt{Gemini-1.5-Flash} (the assessment guidelines, evaluation criteria, and annotator expertise are provided in Appendix~\ref{human_dataset}). The average ratings for each model were as follows:

\begin{itemize}
    \item \textbf{LLaMA-2-7B:} PR: 2.8, BE: 2.7, OF: 2.9, E: 2.6, C: 2.7, H: 2.8, F: 2.9
    \item \textbf{Mistral-7B:} PR: 3.0, BE: 2.9, OF: 3.1, E: 2.8, C: 2.9, H: 3.0, F: 3.1
    \item \textbf{Gemini-1.5-Flash:} PR: 3.5, BE: 3.4, OF: 3.6, E: 3.3, C: 3.4, H: 3.5, F: 3.6
\end{itemize}

Since \texttt{Gemini-1.5-Flash} consistently outperformed the other models across all key dimensions, we selected it for our dataset generation.


\section{Dataset Assessment Details}
\label{human_dataset}

\subsection{Evaluation Aspects of FED and G-Eval Metrics}
For FED metrics, the 18 aspects of dialog qaulity, \textit{viz.} \textit{Interesting, Engaging, Specific, Relevant, Correct, SemanticallyAppropriate, Understandable, Fluent, Overall, Dialog-Level Coherent, Error Recovery, Consistent, Diverse, Topic Depth, Likeable, Understanding, Flexible, Informative,} and \textit{Inquisitive}. For G-Eval, 4 aspects of dialog quality, \textit{viz.} \textit{Engagingness, Naturalness, Coherence,} and \textit{Groundedness} are evaluated.

\subsection{Human Guidelines and Questionnaires}
Before beginning the evaluation, all evaluators undergo a comprehensive briefing on the distinct personality profiles employed in the dialogues. To further enrich their understanding of negotiation dynamics, they are required to watch a detailed video tutorial \cite{chawla2021casino} that showcases a negotiation scenario between two art collectors. The evaluators are provided with metrics description (outlined in Table \ref{manual_metrics}) along with the annotated sample evaluations of the generated conversations to clarify the assessment criteria and rating process. Finally, to preserve objectivity and prevent bias, they are explicitly instructed to conduct the assessments independently, without discussing their ratings or observations with one another at any point during the evaluation. Once trained, the evaluators proceed to rate the conversations using the Likert scale of 1-4, based on the following questions:

\begin{enumerate}[leftmargin=*,topsep=0pt]
\setlength{\itemsep}{0pt}
\setlength{\parskip}{0pt}
\setlength{\parsep}{0pt}
    \item \textbf{Personality Relevance:} 
\textit{To what extent does the generated conversation reflect the traveler's and travel agent's given personalities?}

    Options - \\ 1: Completely irrelevant \\ 2: Mostly irrelevant \\ 3: Mostly relevant \\ 4: Completely relevant


    \item \textbf{Bargaining Efficacy:} 
    \textit{To what extent the traveler/agent presents offers, justifies their position, and makes concessions or compelling arguments that impact the other party’s decisions in the negotiation.}

    Options - \\ 1: Completely ineffective \\ 2: Mostly ineffective \\ 3: Mostly effective \\ 4: Completely effective

    \item \textbf{Outcome Fairness} 
    \textit{To what extent does the final negotiation outcome reflect fairness and equity, resulting in a win-win scenario for both parties?}

    Options - \\ 1: Entirely unfair \\ 2: Mostly unfair \\ 3: Mostly fair \\ 4: Entirely fair


    \item \textbf{Engagineness} 
    \textit{How engaging and content-rich is the overall conversation?} 

    Options - \\ 1: Not at all engaging \\ 2: Somewhat engaging \\ 3: Mostly engaging \\ 4: Very engaging
    

    \item \textbf{Coherent} 
    \textit{How well-structured and logical is the overall conversation?} 
     
     Options - \\ 1: Completely incoherent \\ 2: Mostly incoherent \\ 3: Mostly coherent \\ 4: Completely coherent

    \item \textbf{Humanness} 
    \textit{ Do you think this conversation is from the LLM or a human?} 
    
    Options - \\ 1: Definitely a model \\ 2: Probably a model \\ 3: Probably a human \\ 4: Definitely a human

    \item \textbf{Fluency} 
    \textit{Is the conversation seems contextually natural? Are you able to comprehend the conversation clearly?} 
    
    Options - \\ 1: Very unnatural \\ 2: Mostly unnatural \\ 3: Mostly natural \\ 4: Very natural

\end{enumerate}

\subsection{Human Evaluators Expertise}
All four evaluators possess advanced expertise in dialogue systems and negotiation concepts, with strong proficiency in the English language. Two of the evaluators hold Ph.D. degrees in Linguistics, while the other two have postgraduate degrees in Computer Science, ensuring a well-rounded understanding of both language and computational aspects of the task. Their compensation is provided in accordance with institutional policies.


\begin{table*}[!hbt]
\centering
\begin{adjustbox}{max width=\linewidth}
\begin{tabular}{p{5.8cm}|
                >{\centering\arraybackslash}p{2.9cm}|
                >{\centering\arraybackslash}p{1.5cm}|
                >{\centering\arraybackslash}p{1.5cm}|
                >{\centering\arraybackslash}p{1.5cm}|
                >{\centering\arraybackslash}p{2.4cm}|
                >{\centering\arraybackslash}p{1.5cm}}
\hline
\textbf{Dataset} & \textbf{Domain}  & \textbf{\# Negotiation Setup} & \textbf{\# Conversations} & \textbf{\# of turns} & \textbf{Avg. turns/ conversation} & \textbf{Argume- ntation} 
\\ \hline
STAC \cite{asher2016discourse} & Strategy Games & C & 1,081 & 9,188 & 8.5 & \ding{55}
\\ \hline
DealorNoDeal \cite{lewis2017deal} & Item Assignment & C & 5,808 & 38,332 & 6.6 & \ding{55}   
\\ \hline
CraigslistBargain \cite{he2018decoupling} & E-commerce & NC & 6,682 & 61,474 & 9.2  & \ding{55} 
\\ \hline
NegoCoach \cite{zhou2019dynamic} & Product Bargaining & NC & 300 & - & - & \ding{55} 
\\ \hline
Anti-Scam \cite{li2020end}& E-commerce & NC &  220 & 2,739 & 12.45 & \ding{55} 
\\ \hline
CaSiNo \cite{chawla2021casino} & Item Assignment & C & 1,030 & 11,948 & 11.6 & \ding{55} 
\\ \hline
JobInterview \cite{yamaguchi2021dialogue} & Job Interview & C & 2,639 & 33,515 & 12.7 & \ding{55} 
\\ \hline
\textsc{IND} \cite{ahmad2023ina} & E-commerce & C & 4,163 & 57,393 & 13.79 & \ding{55} 
\\ \hline
\texttt{\textbf{PACT}} \textbf{\textit{(Proposed)}} & Tourism & C & 8,687 & 186,241 & 21.43 & \checkmark 
\\ \hline
\end{tabular}
\end{adjustbox}
\caption{\footnotesize Comparison of \texttt{\textbf{PACT}} with other negotiation conversational corpora. All benchmarks include two-party negotiation. C: Collaborative and NC: Non-collaborative}
\label{compare_datasets}
\end{table*}

\begin{table}[!hbt]
\centering
\begin{adjustbox}{max width=\linewidth}
\begin{tabular}{lccccc} 
\hline
& \textbf{GCQE} & \textbf{PCE} & \textbf{NEE} & \textbf{AEE} & \textbf{TE} \\ \hline
\begin{tabular}[c]{@{}l@{}}Cumulative\\ Survival Rate (\%)\end{tabular} & 87.6 & 80.5 & 71.0  & 67.3 & 63.9 \\
\# of Conversations & 11,894 & 10,934 & 9,645 & 9,138 & 8,687 \\\hline
\end{tabular}
\end{adjustbox}
\caption{\footnotesize Cumulative survival rate with number of conversations in \texttt{\textbf{PACT}} after each filtering step.}
\label{filtering_stats}
\end{table}

\section{Dataset Analysis}\label{dataset_analysis}

\subsection{Diversity and Coverage Analysis}
\begin{figure}[hbt!]
\centering
    \includegraphics[width = \linewidth, scale=1.0]{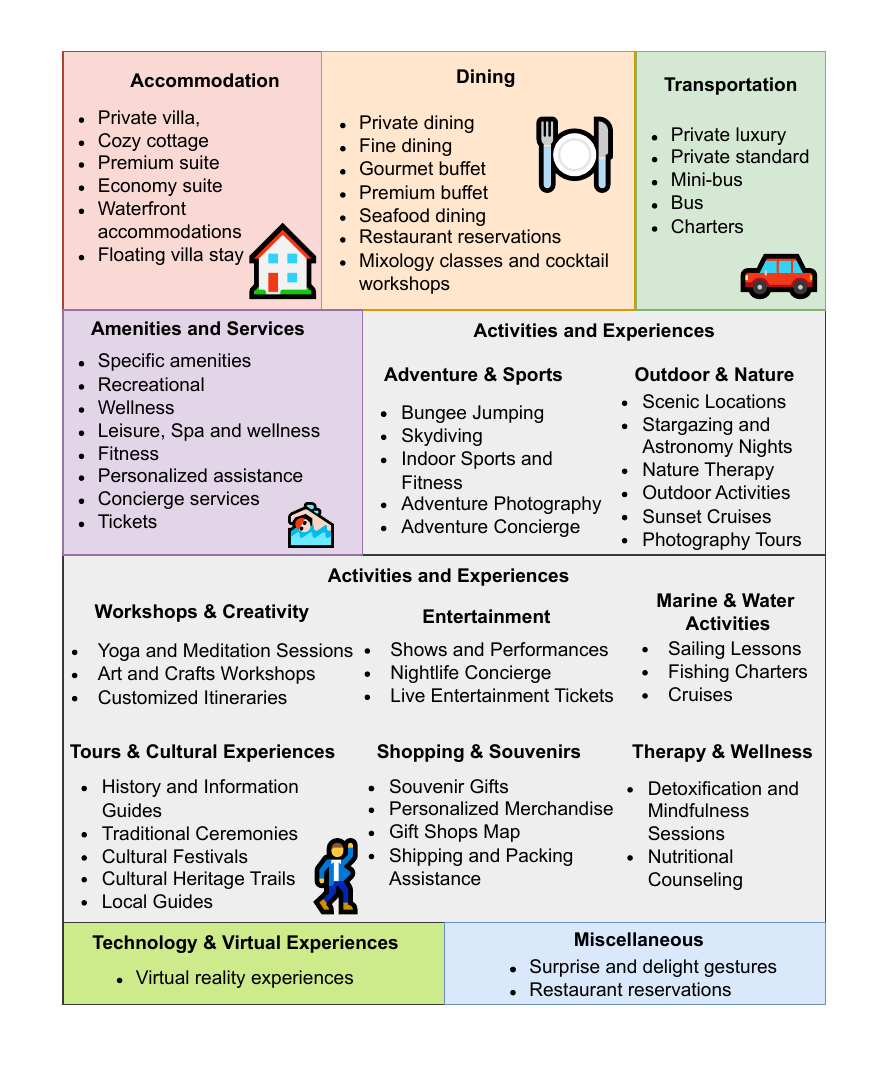}
    \caption{Diversity and coverage of negotiation aspects.}
    \label{pact_diversity}
\end{figure}

We assess the diversity and coverage of negotiation aspects in the \texttt{\textbf{PACT}} dataset through a detailed word frequency analysis. As illustrated in Figure \ref{pact_diversity}, the dataset exhibits significant topic diversity and comprehensive coverage of negotiation elements. It encompasses a broad spectrum of tourism package components, including accommodation, dining, transportation, amenities and services, activities and experiences, technology and virtual experiences, and miscellaneous offerings. This extensive representation ensures that the \texttt{\textbf{PACT}} dataset accurately captures realistic and multifaceted negotiation scenarios, making it a robust resource for research in negotiation dialogue systems.

\subsection{Filtered Dialogue Analysis}
Table \ref{filtering_stats} provides a breakdown of conversation distribution across five filtering steps performed by designated experts. The first step shows a significant reduction, highlighting the general conversation quality expert's role in refining the dataset, with \textasciitilde12\% of initial conversations excluded for not meeting quality standards. This indicates that LLM occasionally fails to generate dialogues with essential qualities. A challenge during personality filtering arises when both traveler and travel agent share similar traits, leading to misclassifications; for example, a negotiator with slightly weaker disagreeableness may be labeled as disagreeable, while another with slightly stronger traits is labeled as argumentative. This tendency to rely on relative rather than absolute traits results in inconsistencies in personality-based filtering. The negotiation and argumentation filtering stages further reduce the number of retained conversations, underscoring the experts' significance in enhancing dataset quality. Also, \textasciitilde36\% of conversations are flagged and removed during toxicity filtering due to harmful or offensive content, a reduction facilitated by the LLM's built-in toxicity mechanisms.

\subsection{Comparison with Existing Datasets}
We present a comparison of our proposed \texttt{\textbf{PACT}} dataset with existing publicly available negotiation dialogue datasets in Table \ref{compare_datasets}. Notably, none of the existing datasets leverages an argumentation-based approach, which we introduce in \texttt{\textbf{PACT}} to enable more rational and realistic negotiation dynamics. Besides, \texttt{\textbf{PACT}} dataset exhibits an average of 21.43 turns per conversation, reflecting a more interactive and conversational structure. It also stands as the largest dataset dedicated exclusively to negotiation scenarios within the tourism domain, further enhancing its relevance for specialized applications.

\section{Experimental Specifications}
\label{experiments}

\subsection{Task Setups Details}
\label{task_setup}
The details of 5 setups are as follows: (i) Pre-trained Models (PMs) to assess LLMs' inherent capability across all sub-tasks; (ii) Pre-trained Models with Personality (PMs\textit{w}P) to examine LLMs' ability to generate personality-driven responses by leveraging the interlocutors' personality as input for RG task; (iii) Fine-tuning on a Human-Human Negotiation Dialogue Dataset (Details are given Appendix \ref{task_setup_details}) (FT-\textit{H2HNego}) investigates whether fine-tuning on general negotiation dialogues improves performance for PR, DAP, and RG tasks; (iv) Fine-tuning on \texttt{\textbf{PACT}} (FT-\textit{Vanilla}) to evaluate the \texttt{\textbf{PACT}}'s effectiveness in improving LLMs performance across all sub-tasks; and (v) Fine-tuning on \texttt{\textbf{PACT}} with background knowledge, personality, and dialog act information (FT-\textit{Ours}) to enhance LLMs' ability to generate factually grounded, personality-driven, and appropriate responses during negotiation. The prompt templates for experiments with GPT-4.1-mini are given in Table \ref{prompt_gpt4_1}. 

\subsection{Response Generation Details}
\label{response_generation}

\subsubsection{Retrieval-Augmented Generation}
We design our response generation framework based on the Retrieval-Augmented Generation (RAG) model \cite{lewis2020retrieval} to integrate background knowledge into LLMs and mitigate the issue of hallucination during RG task. We utilize the curated database of travel packages to collect the package details. To enable efficient retrieval, we generate dense vector representations of these details and the context using BERT-based embeddings \cite{devlin2019bert}. Then, we retrieve the top-most package details that closely align with the dialogue context, ensuring that responses remain coherent and factually accurate. For selecting top-most package details, we then compute the cosine similarity between the dialogue context embeddings and the package embeddings stored in the vector database to retrieve the package exhibiting the highest similarity to the target context. Afterward, we extract the available amenities, services, and their corresponding prices from the retrieved package and incorporate them into the generated responses, thereby ensuring the factual accuracy and reducing the likelihood of exaggerated or misleading descriptions. If a requested amenity or service is unavailable in the retrieved package, we first check for similar alternatives within the same package. If no suitable options are found, the traveler is informed about the unavailability and offered alternative packages to ensure accurate and relevant interaction. 


Once the most relevant package is retrieved, we extract the amenities, services, and their corresponding prices to gather the necessary information for response generation. If a requested amenity or service is unavailable in the retrieved package, our framework incorporates a fallback mechanism to ensure relevant alternatives are identified. Specifically, we search within the same package for amenities or services that are conceptually similar, leveraging semantic similarity using cosine similarity measure to identify the closest match. For creating the embeddings of amenities and services, we again employ BERT model. If no suitable alternatives are found, the agent politely informs the traveler of the unavailability and suggest other packages that might meet the traveler's requirements. This fallback mechanism further ensures that the response remains accurate and relevant. 

\subsubsection{Multi-task Learning}
Acquiring personality information and inferring the future action to be taken from the dialogue context are critical for providing personalized and contextually relevant travel recommendations in real-time. To ensure this, we propose to capture the personality and dialog act information on-the-fly from the dialogue context, which are then used aa a signal for enhanced response generation. The sub-tasks, PR, DAP, and RG tasks are inherently interconnected, as they all require a deep understanding of the dialogue context. To leverage this interdependence and ensure more accurate and consistent response generation, we propose adopting a multi-task learning (MTL) approach \cite{argyriou2006multi} by training the PR, DAP, and RG models simultaneously. The rationale behind this multi-task setup is rooted in the shared nature of the dialogue context input. Each of these models processes the same context and outputs a latent vector, which captures key aspects of the conversation. By training the models together, the model can learn more robust and generalized features that enhance its performance across all three sub-tasks.



\subsection{Human-Human Negotiation Dialogue Dataset Details}
\label{task_setup_details}
For the task setup - `Fine-tuning on a Human-Human Negotiation Dialogue Dataset (FT-\textit{H2Hnego})', we use the Job Interview (JI) dataset that consists of human-human negotiation dialogues in a job interview setting, where participants assume the roles of recruiter and applicant to negotiate distinct issues. The dialogues were collected using Amazon Mechanical Turk (MTurk), ensuring high-quality human-human negotiation interactions. Each negotiation session required workers to exchange at least six messages while making up to three draft agreement proposals, ensuring rich and structured negotiation discourse. The dataset consists of 2,639 dialogues with utterances annotated with one of the dialog acts, namely greet, disagree, agree, inquire, propose, and inform. 

Since the dataset is not annotated with personality profiles, we annotate the dialogues in JI dataset with Big-Five personality traits \cite{mccrae1992introduction} given its universality and wide applicability. Based on the recent advancements in leveraging the LLMs as annotators to significantly reduce annotation costs and human effort \cite{gilardi2023chatgpt,he2023annollm}, we utilize the LLM - LLaMA-3.1-8B \cite{touvron2023llama} due to its strong reasoning capability followed by a human-in-the-loop approach for the interlocutors' profile annotation in the dataset. The entire annotation process unfolds in three steps. 

\begin{enumerate}

\item \textbf{Manual annotation of dialogue samples:} It involves randomly sampling 100 dialogues from the dataset and manually annotating the dialogues with recruiter's and applicant's personality profile labels and the corresponding exaplanations. This annotated dataset is referred to as \textsc{ann-JI}, hereafter.

\item \textbf{Few-shot prompting for personality profile labeling:} In this step, LLaMA-3.1-8B \cite{touvron2023llama} is prompted in a few-shot setting, using the manually annotated samples as few-shot demonstrations, to generate personality profile labels. These labels are accompanied by natural language explanations to provide insight into the model’s predictions. The sample prompt is shown below. 

\begin{tcolorbox}[title=Prompt template for personality profile labeling]
\small
\textbf{Task instruction:} Given the dialogue, determine the recruiter's and applicant's personality as one of the following - \textit{Openness}, \textit{Conscientiousness}, \textit{Extroversion}, \textit{Agreeableness}, and \textit{Neuroticism}. Then explain your prediction. 

\textbf{Few-shot demonstrations:} 

Dialogue:

Recruiter's Personality Label:

Applicant's Personality Label:

Explanation: 

\textbf{Target input:}

Dialogue:

\textbf{LLM output:}

Recruiter's Personality Label: $<$label1$>$

Applicant's Personality Label: $<$label2$>$

Explanation: $<$explanation$>$

\end{tcolorbox}

\item \textbf{Manual Verification and Correction:} It involves verifying and re-annotating the dialogues by the human subjects to create gold-standard datasets. \end{enumerate}

\noindent \textbf{Annotation Guidelines. }Each dialogue is annotated by 3 annotators. The annotators are instructed to analyze the overall interaction style of the interlocutors rather than focusing on individual utterances. If a speaker exhibits multiple traits throughout the conversation, the most dominant one should be selected to capture the interlocutor's general personality rather than momentary shifts in behavior. To enhance reliability, we employ a majority voting mechanism to determine the final personality profile of both the applicant and the recruiter. We observe a reliable multi-rater Kappa \cite{mchugh2012interrater} agreement ratios of 79.2\% in Step 1 and 81.8\% Step 3 among the annotators. It is noted that dataset annotation involves the same human evaluators as those involved in manual assessment of \texttt{\textbf{PACT}} dataset quality (\textsection \ref{PACT_quality_assessment}).

\subsection{Implementation Details}
We split the dataset into train, validation, and test with the ratio of 7:1:2. All the results are computed on the test set. We prompt the LLMs in zero-shot setup. For conversation generation, we adopt nucleus (Top-$p$) sampling \citep{holtzman2019curious} with $p = 0.95$ and temperature $T = 1.0$. We fine-tune the LLMs with LoRA adaption \cite{hu2021lora} using 4*V100 GPUs, with batch size of 2 per GPU, and learning rate of $5e^{-5}$ for three epochs. The fine-tuning of all the LLMs are based on the SFT trainer of Transformer Reinforcement Learning (TRL)\footnote{\url{https://github.com/huggingface/trl}} \cite{von2020trl}. All the implementation is done using the PyTorch\footnote{\url{https://pytorch.org/}} framework. The transformer-based models from HuggingFace \cite{wolf2019huggingface} are utilized in all the experiments.

\subsection{Evaluation Metrics Details}
\label{metrics_details}

Perplexity (PPL) \cite{papineni2002bleu} evaluates how well a model predicts a response. Word-overlap-based metrics like BLEU-1/2 (B-1/2) \cite{papineni2002bleu}, METEOR (MET) \cite{banerjee2005meteor}, and ROUGE-L (R-L) \cite{lin2004rouge} compute the n-gram overlap between the ground-truth response and the model's generated response. DISTINCT-1/2 (D-1/2) \cite{li2015diversity} measures the diversity of n-grams in the generated responses. Embedding-based metrics such as BERTScore F1 (BS-F1)\footnote{BERTScore: \url{https://huggingface.co/spaces/evaluate-metric/bertscore}} \cite{zhang2019bertscore} align the generated response and the ground-truth response in latent semantic space to assess the semantic similarity between the gold response and the model's generated response. BS-F1 computes word similarity using contextual embeddings from the BERT \cite{devlin2019bert} model. 

Response Length (R-LEN) calculates the length of the generated responses, Personality Accuracy (P-ACC) evaluates the alignment of the generated responses with the specified personality profile, and Dialog Act Accuracy (D-ACC) measures the consistency of the generated responses with the designated dialog act. To evaluate P-ACC and D-ACC, we fine-tune two distinct RoBERTa-base models \cite{liu2019roberta} on our dataset for 8 epochs. The classifier used for computing P-ACC achieves an accuracy of 61.57\% and a macro-F1 score of 58.31\%. Similarly, the classifier for D-ACC attains an accuracy of 80.59\% and a macro-F1 score of 79.22\%.

FEQA \cite{durmus2020feqa} and QuestEval (QE) \cite{scialom2021questeval} are QA-based metrics designed to evaluate the faithfulness of generated content. FEQA assesses consistency by extracting answers from a source document using question-answer pairs derived from the summary, with discrepancies indicating unfaithfulness. It demonstrates strong alignment with human judgments and excels in evaluating highly abstractive summaries. In this study, we adapt FEQA by concatenating the dialogue context and grounded package information to form the document and treating the generated response as the summary. QE operates in both reference-dependent and reference-free modes, with the former relying on input sources and golden references. In this study, the reference-dependent mode is used, employing the same source construction method as FEQA. Both FEQA\footnote{FEQA: \url{https://github.com/esdurmus/feqa}} and QE\footnote{QE: \url{https://github.com/ThomasScialom/QuestEval}} metrics are computed using their respective official libraries with default settings.

\section{Human Evaluation for the Response Generation Task}
\label{human_eval}

\subsection{Evaluation Metrics and Instructions}
Human evaluation for the RG task is conducted by the same four evaluators involved in the dataset creation process. They assess 50 randomly selected responses based on six criteria: Personality Relevance (PR), Bargaining Efficacy (BE), Outcome Fairness (OF), Fluency (F), Coherence (C), and Engagingness (E), using a 1-4 scale (low to high). Further, they assess the faithfulness of the generated responses (evaluating whether the responses adhere to the to the dialogue context and grounded package information) using Faithful (Faith.) and Hallucinated (Hal.). For each sample, evaluators are provided with the dialogue context, ground truth response, and the LLM-generated response to rate each response across these metrics adhering to the instructions mentioned below .

\begin{enumerate}[leftmargin=*,topsep=0pt]
\setlength{\itemsep}{0pt}
\setlength{\parskip}{0pt}
\setlength{\parsep}{0pt}
    \item \textbf{Personality Relevance (PR)} evaluates whether the generated response aligns with the established personality profile. 

    Instruction: \textit{To what extent does the generated response reflect the established personality profile of the traveler and travel agent?}

    Scale: 1 (not at all aligned with the personality profile) to 4 (perfectly aligned with the personality profile).

    \item \textbf{Bargaining Efficacy (BE)} measures the effectiveness of the generated response in presenting compelling offers and arguments that influence the other party's decisions, ultimately achieving favorable outcomes.

    Instruction: \textit{How effectively does the generated response present persuasive offers and arguments during the negotiation?}

    Scale: 1 (ineffective and unpersuasive) to 4 (highly effective and persuasive).

    \item \textbf{Outcome Fairness (OF)} evaluates the equity of the negotiation results by analyzing how well the generated response reflects an agreement that meets the interests of both the traveler and the agent. 

    Instruction: \textit{To what extent does the generated response indicate a fair and equitable agreement between the parties?} 
    
    Scale: 1 (entirely unfair, heavily favoring one party) to 4 (perfectly equitable, achieving a balanced win-win outcome).

    \item \textbf{Fluency (F)} evaluates the grammatical correctness of the generated response.

     Instruction: \textit{How grammatically correct and fluent is the generated response?} 
     
     Scale: 1 (severely flawed with numerous errors) to 4 (fluent and grammatically correct).

    \item \textbf{Coherence (C)} examines how well the generated responses align with the dialogue context. 
    
    Instruction: \textit{To what extent does the generated response maintain logical flow and relevance within the conversation?} 
    
    Scale: 1 (completely incoherent and irrelevant) to 4 (perfectly coherent and contextually relevant).

    \item \textbf{Engagingness (E)} measures the extent to which the generated response is engaging, compelling, and capable of retaining users' attention.

    Instruction: \textit{How captivating and interesting is the generated response in retaining the user's attention?} 
    
    Scale: 1 (not engaging at all) to 4 (extremely engaging and captivating).

    \item \textbf{Faithfulness (Faith.) and Hallucinated (Hal.)} evaluate whether the response is faithful or hallucinated. 

     Instruction: \textit{Whether the generated response is faithful or hallucinated? A response is faithful if it aligns with the dialogue context and package information, whereas it is labeled as hallucinated if it either contradicts the provided sources or lacks corroboration from them.}
    
    Please choose one of the following options:
    
    A. The generated response is faithful.
    
    B. The generated response contains intrinsic hallucination only.
    
    C. The generated response contains extrinsic hallucination only.
    
    D. The generated response contains both intrinsic and extrinsic hallucinations.
    
    \textbf{Enter Your Option (Only enter A or B or C or D):} 

\end{enumerate}

\subsection{Evaluation Results}
Table \ref{human_res_rg} presents the results of human evaluation. We conduct human evaluation for FT-\textit{Vanilla} and FT-\textit{Ours} setups only, as manual evaluation is expensive. From the results, we observe that the findings closely align with automatic evaluation results, providing a comprehensive assessment of model performance under FT-\textit{Vanilla} and FT-\textit{Ours} setups. In the FT-\textit{Vanilla} setup, Vicuna achieves the highest scores across all key metrics - PR, BE, OF, F, C, and E, with values of 3.52, 3.45, 3.17, 3.39, 3.41, and 3.55, respectively. This mirrors its strong performance in automatic evaluations, reinforcing its ability to generate fluent, coherent, and engaging responses during negotiation. Additionally, Llama-2-chat and Llama-3.1 demonstrate competitive performance, particularly in fluency (F = 3.56, 3.67) and coherence (C = 3.28, 3.01), indicating their effectiveness in maintaining contextual relevance. However, models trained under the FT-\textit{Ours} setup with background package, personality and dialog act information exhibit substantial improvements across all evaluation dimensions. Notably, Vicuna outperforms all models with PR = 3.92, BE = 3.81, OF = 3.91, F = 3.93, C = 3.82, and E = 3.79, showing consistent gains over FT-\textit{Vanilla}. These results highlight the effectiveness of our fine-tuning approach with background package, personality and dialog act information in enhancing response personalization, negotiation efficacy, coherency, fluency, and engagingness. Compared to its FT-\textit{Vanilla} counterpart, Vicuna achieves absolute improvements of 10.20\%, 9.45\%, 18.93\%, 13.74\%, 10.73\%, and 6.33\% across PR, BE, OF, F, C, and E, respectively.

The faithfulness (Faith.) and hallucination (Hal.) metrics further underscore the reliability of models under FT-\textit{Ours}. Across all models, Faith. scores increase significantly, with Vicuna achieving the highest Faith. score of 77.98 and the lowest Hal. score of 22.02. Further, the proposed FT-Ours setup outperforms ChatGPT across all metrics for the RG task. These results indicate that our fine-tuning method not only enhances content quality but also improves factual accuracy and reduces hallucinations, making negotiation responses more trustworthy and consistent. Overall, the results affirm that our fine-tuning approach (FT-\textit{Ours}) significantly enhances response quality across personalization, coherence, negotiation efficacy, and faithfulness. The substantial improvements across all models, particularly for Vicuna highlight the impact of our approach in refining dialogue generation for negotiation scenarios. To assess the reliability of human evaluations, Fleiss's Kappa \cite{mchugh2012interrater} is computed to measure the inter-evaluator agreement. The average Kappa scores of 0.83, 0.83, 0.80, 0.78, 0.77, 0.85, 0.83 are obtained for PR, BE, OF, F, C, E, Faith. (or Hal.), respectively, implying substantial agreement among the evaluators. 

\begin{table}[!hbt]
\centering
\begin{adjustbox}{max width=\linewidth}
\begin{tabular}{c|l|cccccccc} \hline
\multicolumn{1}{l|}{\textbf{Task Setups}} & \textbf{Models} & \textbf{PR} & \textbf{BE} & \textbf{OF} & \textbf{F} & \textbf{C} & \textbf{E} & \textbf{Faith.} & \textbf{Hal.} \\ \hline
\multirow{6}{*}{\begin{tabular}[c]{@{}c@{}}FT-\textit{Vanilla} \end{tabular}}                
                            & Llama-2       & 3.11 & 2.94 & 3.07 & 3.12 & 2.84 & 2.99 & 54.32 & 45.68\\ 
                            & Llama-2-chat  & 3.34 & 3.34 & 3.15 & 3.56 & 3.28 & 2.94 & 58.46 & 41.54 \\ 
                            & Llama-3       & 2.73 & 3.28 & 2.99 & 3.19 & 3.02 & 2.65 & 53.27 & 46.73\\ 
                            & Llama-3.1     & 3.45 & 3.11 & 2.86 & 3.67 & 3.01 & 3.43 & 60.14 & 39.86\\ 
                            & Mistral       & 2.86 & 3.03 & 3.08 & 2.91 & 3.17 & 3.29 & 59.88 & 40.12 \\ 
                            & Vicuna        & 3.52 & 3.45 & 3.17 & 3.39 & 3.41 & 3.55 & 63.42 & 36.58\\ \hline
\multirow{6}{*}{FT-\textit{Ours}} & Llama-2       & 3.51 & 3.49 & 3.22 & 3.55 & 3.42 & 3.61 & 75.31 & 24.69 \\
                               & Llama-2-chat  & 3.89 & 3.69 & 3.75 & 3.31 & 3.66 & 3.72 & 73.81 & 26.19 \\
                               & Llama-3       & 3.55 & 3.61 & 3.39 & 3.79 & 3.55 & 3.41 & 76.17 & 23.83 \\
                               & Llama-3.1     & 3.74 & 3.58 & 3.63 & 3.82 & 3.45 & 3.56 & 71.58 & 28.42 \\
                               & Mistral       & 3.64 & 3.44 & 3.65 & 3.35 & 3.42 & 3.53 & 73.19 & 26.81 \\
                               & Vicuna        & \textbf{3.92} & \textbf{3.81} & \textbf{3.91} & \textbf{3.93} & \textbf{3.82} & \textbf{3.79} & \textbf{77.98 }& \textbf{22.02} \\ \hline
                               & GPT-4.1-mini & 3.34 & 3.28 & 3.15 & 3.29 & 3.17 & 3.22 & 65.34 & 34.66 \\ \hline

\end{tabular}
\end{adjustbox}
\caption{Human evaluation results for response generation task. All results are
statistically significant based on Welch’s t-test \cite{welch1947generalization} conducted at a 5\% significance level. Best scores are in bold.}
\label{human_res_rg}
\end{table}


\subsection{Case Study}
We analyze a few samples from the test set and present the responses generated by the top two LLMs - Vicuna and Llama-2-chat under two task setups, \textit{viz.} FT-\textit{Vanilla} and FT-\textit{Ours} in Table \ref{qual_analysis}. Our observations indicate that integrating backaground package knowledge together with personality and dialog act information generates more factual, rational and convincing responses aligned with the personality and dialog act during negotiation. For instance, in the first sample, traveler expresses concern about the price of a travel package and seeks a reduction. The FT-\textit{Vanilla} models (LLaMA-2-chat, Vicuna) reinforce the price but fail to explore alternatives, making their responses appear rigid and less engaging. Additionally, these models sometimes introduce minor inconsistencies or fail to align completely with the negotiation context. In contrast, FT-\textit{Ours} models not only recognize the traveler’s budget constraints but also proactively propose factual modifications, such as adjusting accommodations or tweaking the itinerary. The responses remain grounded in the provided package information, ensuring that no hallucinated details are introduced while maintaining persuasiveness, flexibility, and user engagement. In the second sample, traveler requests the removal of a package component to lower the overall cost. Here, the ground-truth agent provides a price adjustment while maintaining the overall value of the package. The FT-\textit{Vanilla} models largely mirror this response but occasionally generate exaggerated or incorrect details, such as an arbitrary price increase. However, the FT-\textit{Ours} models demonstrate a more adaptive negotiation action while maintaining factual correctness. They suggest reasonable itinerary adjustments to better match the traveler’s budget while ensuring quality retention, all without introducing fabricated details.

Overall, the analysis demonstrates that FT-\textit{Ours} consistently outperforms other setups by generating responses that are both effective and factually grounded. By leveraging background knowledge, personality profiles, and dialog act cues, FT-\textit{Ours} models enhance negotiation adaptability, improve persuasiveness, and foster greater context-awareness while minimizing hallucinations. These findings underscore the importance of incorporating package knowledge, personality, and dialog act information into LLMs for more effective, truthful, and human-like interactions.

Moreover, we conduct human evaluation in a multi-turn setup for the best-performing Vicuna model under FT-Ours setup. The human evaluation is conducted with the help of the same four human evaluators. Before interacting with the system, the evaluators receive a detailed briefing on the travel packages used in the dataset, including associated aspects, amenities, and service descriptions, followed by the different personality profiles. Evaluators are instructed to select a particular travel package and one profile each from argumentation, preference, and buying style profile sets, and initiate conversations with the system based on a given travel package and express their preferences according to the selected personality profiles.

Each evaluator engages in 5 interactions, ensuring that every session features a different travel package with a distinct combination of attributes and system responses. To ensure meaningful dialogue, each interaction spans at least 8 conversational turns. This setup results in a total of 20 human-evaluated dialogues. After completing all interactions, evaluators rate each dialogue along six dimensions: PR, BE, OF, F, C, E, Faith, and Hal. using a 4-point Likert scale. We obtain the following results: PR: 3.05, BE: 3.92, OF: 3.94, F: 3.01, C: 3.87, E: 3.85, Faith.: 79.16, Hal.: 20.84. These high ratings across all metrics demonstrate that Vicuna, under the FT-Ours setup, consistently generates personality-relevant responses, facilitates effective bargaining, and ensures fairness in negotiation outcomes. Besides, responses are consistently rated as fluent, coherent, natural, and faithful, underscoring the overall high quality of the generated negotiation interactions. A sample interaction is given in Table \ref{casestudy_multiturn}.


\begin{table*}[!hbt]
\centering
\begin{adjustbox}{max width=\linewidth}
\begin{tabular}{p{20cm}}
\hline
\begin{verbatim}
{
    "packageX": {
        "Travel_Package_Name": "Sight Tour",
        "Description": "Always ready to stop for that landmark, event, scenic vista or attraction",
        "Services": {
            "accommodation": {
                "private villa": 25000,
                "cozy cottage": 15000,
                "premium suite": 8000,
                "economy suite": 4000
            },
            "meals": {
                "private dining": 15000,
                "fine dining": 9000,
                "gourmet buffet": 6000,
                "premium buffet": 3000
            },
            "transportation": {
                "private luxury": 30000,
                "private standard": 15000,
                "mini-bus": 8000,
                "bus": 5000
            }
        },
        "Optional_amenities": [
            "specific",
            "recreational",
            "wellness",
            "leisure",
            "personalized"
        ],
        "options": [
            [
                ["history and information guides", 521],
                ["indoor sports and fitness", 669],
                ["spa and wellness facilities", 732],
                ["reading retreats", 482],
                ["customized itineraries", 787]
            ],
            [
                ["souvenir gifts", 2218],
                ["game room", 1960],
                ["yoga and meditation sessions", 2277],
                ["art and crafts workshop", 2008],
                ["surprise and delight gestures", 2491]
            ],
            [
                ["local guides", 3404],
                ["entertainment shows and performances", 3437],
                ["nutritional counselling", 3233],
                ["stargazing and astronomy nights", 3023],
                ["celebratory packages", 3670]
            ],
            [
                ["virtual reality experiences", 4932],
                ["outdoor activities", 4943],
                ["detoxification and mindfulness session", 4913],
                ["photography tour", 4587],
                ["personal assistance", 5162]
            ]
        ]
    }
} 
\end{verbatim} \\\hline

\end{tabular}
\end{adjustbox}
\caption{Sample travel package information.}
\label{package_info}
\end{table*}

\begin{table*}[!hbt]
\centering
\begin{adjustbox}{max width=\linewidth}
\begin{tabular}{p{20cm}}
\hline
\textbf{General Conversation Quality Expert} \\ \hline 
    
\textcolor{blue}{\texttt{Coherence}}
    
\textcolor{purple}{\texttt{[Instruction]}} \texttt{Is the conversation coherent, and does it maintain a smooth, consistent flow from start to finish? Please rate the \textit{coherence} of the given conversation on a scale of 1 to 3, where 1 represents low coherence and 3 indicates a high level of coherence. Also, provide the rationale for your rating.} 
    
\textbf{Input} 
    
\{Conversation\}
    
\textbf{Output}
    
Rating: 
    
Rationale:  \\\hline

\textcolor{blue}{\texttt{Consistency}}
    
\textcolor{purple}{\texttt{[Instruction]}} \texttt{Does the conversation maintain consistency in the information presented throughout its entirety? Please rate the \textit{consistency} of the given conversation on a scale of 0 to 1, where 0 signifies inconsistency and 1 indicates strong consistency. Also, provide the rationale for your rating.}
    
\textbf{Input} 
    
\{Conversation\}
    
\textbf{Output}
    
Rating: 
    
Rationale:   \\\hline

\textcolor{blue}{\texttt{Diversity}}
    
\textcolor{purple}{\texttt{[Instruction]}} \texttt{Is there diversity in the responses throughout the conversation? Please rate the \textit{diversity} of the given conversation on a scale of 1 to 3, where 1 signifies low diversity and 1 indicates strong diversity. Also, provide the rationale for your rating.} 
    
\textbf{Input} 
    
\{Conversation\}
    
\textbf{Output}
    
Rating: 
    
Rationale: \\\hline

\textcolor{blue}{\texttt{Topic Depth}}
    
\textcolor{purple}{\texttt{[Instruction]}} \texttt{Does the conversation delve into topics in depth? Please rate the \textit{topic depth} of the given conversation on a scale of 1 to 3, where 1 indicates a superficial discussion and 3 signifies a comprehensive exploration. Also, provide the rationale for your rating.}
    
\textbf{Input} 
    
\{Conversation\}
    
\textbf{Output}
    
Rating: 
    
Rationale:  \\\hline

\textcolor{blue}{\texttt{Understanding}}
    
\textcolor{purple}{\texttt{[Instruction]}} \texttt{Does the conversation demonstrate an understanding of the user? Please rate the understanding of the given conversation on a scale of 1 to 3, where 1 signifies a lack of understanding and 3 indicates strong understanding. Also, please provide a rationale for your rating.}
    
\textbf{Input} 
    
\{Conversation\}
    
\textbf{Output}
    
Rating: 
    
Rationale:  \\\hline

\textcolor{blue}{\texttt{Flexibility}}
    
\textcolor{purple}{\texttt{[Instruction]}} \texttt{Is the conversation flexible and adaptable to the user and their interests? Please rate the \textit{flexibility} of the given conversation on a scale of 1 to 3, where 1 signifies low flexibility and 3 indicates high flexibility. Additionally, provide a rationale for your rating.}
    
\textbf{Input} 
    
\{Conversation\}
    
\textbf{Output}
    
Rating: 
    
Rationale:  \\\hline

\textcolor{blue}{\texttt{Likeability}}
    
\textcolor{purple}{\texttt{[Instruction]}} \texttt{Does the conversation exhibit a likable personality? Please rate the \textit{likeability} of the given conversation on a scale of 1 to 3, where 1 indicates an unlikable personality and 3 signifies a highly likable personality. Also, provide the rationale for your rating. }
    
\textbf{Input} 
    
\{Conversation\}
    
\textbf{Output}
    
Rating: 
    
Rationale: \\\hline
\end{tabular}
\end{adjustbox}
\caption{Prompt templates used for General Conversation Quality Expert for \texttt{\textbf{PACT}} filtering. Variables enclosed in curly brackets {} represent placeholders that are filled when the template is instantiated.}
\label{filter_prompt1}
\end{table*}

\begin{table*}[!hbt]
\centering
\begin{adjustbox}{max width=\linewidth}
\begin{tabular}{p{20cm}}
\hline
    
    \textbf{Personality Consistency Expert} \\\hline
    
    \textcolor{purple}{\texttt{[Instruction]}} \texttt{Imagine a travel agent with an argumentation profile as \{argumentation profile of the travel agent with description\} and a traveler with an argumentation profile as \{argumentation profile of the traveler with description\}, preference profile as \{preference profile of the traveler with description\}, and buying style profile as \{buying style profile of the traveler with description\}. Does the conversation align with the established personalities of both the traveler and the travel agent throughout? Please rate the personality consistency on a scale of 1 to 3, where 1 represents poor alignment with the personalities and 3 indicates strong consistency in personality portrayal. Also, provide the rationale for your rating.}
    
    \textbf{Input} 
    
    \{Conversation\}
    
    \textbf{Output}
    
    Rating: 
    
    Rationale: \\ \hline
    \textbf{Negotiation Efficacy Expert} \\\hline
    
    \textcolor{purple}{\texttt{[Instruction]}} \texttt{How effective is the negotiation between the traveler and the travel agent in creating a win-win scenario? Please rate the negotiation efficacy of the conversation on a scale of 1 to 3, where 1 indicates poor effectiveness in achieving mutual agreement and 3 signifies strong effectiveness in fostering a win-win outcome. Also, provide the rationale for your rating.} 
    
    \textbf{Input} 
    
    \{Conversation\}
    
    \textbf{Output}
    
    Rating: 
    
    Rationale: \\ \hline
    \textbf{Argumentation Efficacy Expert}  \\\hline
    \textcolor{purple}{\texttt{[Instruction]}} \texttt{How well do the negotiating parties manage arguments and present persuasive rationales throughout the conversation? Please evaluate the argumentation efficacy on a scale of 1 to 3, where 1 represents poor argumentation skills and 3 indicates strong ability to present convincing arguments for mutually beneficial outcomes. Also, provide the rationale for your rating.}
    
    \textbf{Input} 
    
    \{Conversation\}
    
    \textbf{Output}
    
    Rating: 
    
    Rationale: \\ \hline
    \textbf{Toxicity Expert} \\\hline \textcolor{purple}{\texttt{[Instruction]}} \texttt{Is the conversation toxic? Please rate the \textit{toxicity} of the given conversation on a scale of 0 to 1, where 0 represents non-toxic and 1 indicates toxic. Also, provide the rationale for your rating.} 
    
    \textbf{Input} 
    
    \{Conversation\}
    
    \textbf{Output}
    
    Rating: 
    
    Rationale:  \\\hline
\end{tabular}
\end{adjustbox}
\caption{Prompt templates used for various experts for \texttt{\textbf{PACT}} filtering. Variables enclosed in curly brackets {} represent placeholders that are filled when the template is instantiated.}
\label{filter_prompt2}
\end{table*}

\begin{table*}[!hbt]
\centering
\begin{adjustbox}{max width=\linewidth}
\begin{tabular}{p{18cm}}
\hline
\texttt{\textcolor{purple}{[Dialog Act]}} \texttt{Greet-Ask}

\texttt{\textcolor{purple}{[Task Overview]}} \texttt{You are a travel agent, and you have to greet the traveler and ask for traveler's preference for booking the desired tour package.}

\textbf{Few-shot demonstrations}

\texttt{\textcolor{purple}{[Sample Utterance]}} \texttt{Hello there. Welcome, I am happy to help you choose the travel package according to your preferences. Can you please tell me what kind of travel experience you are planning for yourself?}

\textbf{Input}

\texttt{You are a travel agent, and you have to greet the traveler and ask for traveler's preference for booking the desired tour package.}

\textbf{Output}

$<$\texttt{Generated utterance}$>$

\\\hline

\texttt{\textcolor{purple}{[Dialog Act]}} \texttt{Elicit-preference}

\texttt{\textcolor{purple}{[Task Overview]}} \texttt{You are a human traveler and you are describing the kind of tour package you are planning according to your personality profile. Your personality description is: Vacations, relaxing in quiet and peaceful places. You will say:}

\textbf{Few-shot demonstrations}

\texttt{\textcolor{purple}{[Sample Utterance]}} \texttt{I want to enjoy vacations where I can get away from it all in order to relax in quiet and peaceful places.}

\textbf{Input}

\texttt{You are a human traveler and you are describing the kind of tour package you are planning according to your personality profile. Your personality description is \{description\}. You will say:}

\textbf{Output}

$<$\texttt{Generated utterance}$>$

\\\hline

\texttt{\textcolor{purple}{[Dialog Act]}} \texttt{Inform}

\texttt{\textcolor{purple}{[Task Overview]}} \texttt{You are a travel agent and you are giving information about a package to the traveler. The name of the package is `SightTour'. The cost of the package is 53432\$, the package has cozy cottage as accommodation option, fine dining as meal option and private standard as transportation option. The package has the amenities of virtual reality experiences, detoxification and mindfulness sessions and outdoor activities. You will provide the above mentioned information by replying:}

\textbf{Few-shot demonstrations}

\texttt{\textcolor{purple}{[Sample Utterance]}} \texttt{Great! You will love our `SightTour' package. The package will cost you 53432\$ and you will be getting amazing amenities like virtual reality experiences, detoxification and mindfulness sessions and outdoor activities. You will get cozy cottage as accommodation, fine dining and private standard transportation. Would you like to go with this package?}

\textbf{Input}

\texttt{You are a travel agent and you are giving information about a package to the traveler. The name of the package is \{package\}. The cost of the package is \{cost\}\$, the package has \{service\}. The package has the amenities of \{amenity\}. You will provide the above mentioned information by replying:}

\textbf{Output}

$<$\texttt{Generated utterance}$>$

\\\hline

\end{tabular}
\end{adjustbox}
\caption{Prompt template for different dialog acts that are used for conversation generation. Variables enclosed in curly brackets {} represent placeholders that are filled when the template is instantiated.}
\label{conv_generate_prompt}
\end{table*}

\begin{table*}[!hbt]
\centering
\begin{adjustbox}{max width=\linewidth}
\begin{tabular}{c|c|p{15cm}}
\hline

\multicolumn{3}{l}{\textbf{Traveler Personality Profile:} \{\textcolor{brown}{Disagreeable, Beach Lover, Quality\&Budget-concerned}\} \textbf{Travel Agent Personality Profile:} \{\textcolor{brown}{Open-minded}\}}
\\\hline
\textbf{Interlocutor}  & \textbf{Dialog Act}  & \textbf{Utterance} \\ \hline
Travel Agent & Greet-Ask & Welcome! I'm so excited to help you plan your dream trip. To get started, tell me a little about what you're looking for. Are you interested in a relaxing beach vacation, an adventurous hike through the mountains, a bustling city break, or something else entirely? What kind of budget are you working with, and how long do you have for your trip? \\ 
Traveler & Elicit-Preference & I'm looking for a vacation that's all about soaking up the sun and enjoying the water. Something with plenty of beaches, maybe a bit of snorkeling or paddle boarding, and some delicious seafood restaurants. I'm thinking a tropical paradise with warm, clear waters, white sand beaches, and maybe a few palm trees for good measure! \\ 
Travel Agent & Inform & For the ultimate beach getaway, I recommend our 'Beach Bum' package. It's priced at \$54,225 and includes cozy cottage accommodations, fine dining, and private standard transportation. You'll enjoy snorkeling and diving excursions, glass-bottom boat tours, and private beach access. Sounds like a perfect escape, right? \\ 
Traveler & Ask-clarification-X & Hey there! I'm really interested in doing some snorkeling and diving while I'm here. What kind of excursions do you offer? Are there any beginner-friendly spots? And what kind of marine life can I expect to see? \\ 
Travel Agent & Provide-clarification-X & Our snorkeling and diving excursions are truly spectacular! Imagine yourself gliding through crystal-clear waters, surrounded by vibrant coral reefs teeming with colorful fish. It's an experience that will leave you breathless! We offer options for all levels, from beginner snorkelers to experienced divers. Plus, our expert guides will ensure your safety and point out all the hidden gems beneath the surface. It's a unique way to connect with nature and create memories that will last a lifetime! \\ 
Traveler & Negotiate-price-decrease & I really appreciate the tour package you've put together, but \$54,225 is a bit out of my budget. Would you be willing to work with me on a price closer to \$40,672? I'm confident we can find a solution that works for both of us. \\ 
Travel Agent & Negotiate-price-increase & I appreciate you offering a price, and I understand you're looking for a great deal. However, \$40,672 is quite a bit lower than what's needed to cover the costs of this comprehensive package. To make this work for both of us, I'd propose a revised price of \$51,908.59. This would ensure you receive the high-quality experience we're known for, while still being a competitive rate for the value you're getting. \\ 
Traveler & Negotiate-price-decrease & I appreciate the package you've put together, but unfortunately, my budget for this trip is closer to \$43,193.15. Would you be willing to work with me to see if we can reach a price closer to that figure? \\ 
Travel Agent & Negotiate-price-increase & I appreciate you proposing a price, and I understand you're looking for a great deal. However, to offer you the best possible experience with this package, we need to adjust the price slightly to \$47,550.51. This ensures we can provide you with the top-quality accommodations, excursions, and service you deserve. Let's discuss what's included in the package and how we can customize it to fit your budget and preferences. \\ 
Traveler & Disagree-price & I appreciate you putting together this package for me, but \$47,550.51 is a bit out of my budget. I was hoping to find a package closer to \$45,550. Is there any flexibility in the pricing, perhaps with some slight adjustments to the itinerary? I'm open to suggestions. \\ 
Travel Agent & Justify-price & I understand your concern, but this package offers great value with luxury accommodations, exclusive experiences, flights, meals, and activities. It includes direct Emirates flights, a 5-star beachfront villa, snorkeling tours, dolphin-watching cruises, and gourmet dining. Plus, with our 10\% discount, the price is significantly reduced. Similar packages elsewhere would cost over \$55,000. Let’s discuss your priorities and see if we can adjust the package, perhaps switching the villa for a suite or tweaking activities to fit your budget. What do you think?
 \\ 
Traveler & Disagree-price & While I'm interested in this tour package, \$47,550.51 seems a bit steep. Could we discuss a lower price? I'm sure we can find a figure that works for both of us. \\ 
Travel Agent & Negotiate-add-X & Instead of private beach access, I'm suggesting we add a breathtaking sunset cruise to your package. Imagine enjoying stunning views as the sun dips below the horizon. \\ 
Traveler & Provide-consent & That sounds amazing! I'd love to add the sunset cruises to the package. What time do they usually depart? \\ 
Travel Agent & Consent-response & Great! Sunset cruises are now included in your package, bringing the total cost to \$46,975.51. \\ 
Traveler & Accept & Okay, I'm happy with the details of the tour package. Let's go ahead and book it. \\ 
Travel Agent & Acknowledge-acceptance & Wonderful! We're thrilled you've chosen this tour package. Thank you for your business! \\ \hline
\end{tabular}
\end{adjustbox}
\caption{A sample generated conversation between the traveler and travel agent engaged in argumentation-based negotiation.}
\label{sample_conversation}
\end{table*}


\begin{table*}[hbt!]
\centering
\begin{adjustbox}{max width=\linewidth}
\begin{tabular}{p{20cm}}
\hline
\textbf{Personality Recognition Task} \\ \hline
\texttt{\textcolor{purple}{[Task Instruction]}} \texttt{Imagine you are a travel agent negotiating with a potential traveler using an argumentation-based approach. The focus of the negotiation is on a \{\} travel package described as \{\}, where both parties aim to reach a mutually beneficial agreement. Given the conversation between traveler and travel agent, your task is to predict the argumentation profile, preference profile, and buying style profile of the traveler and argumentation profile of the travel agent. Argumentation profile of the traveler can be one of the following - \{list\_of\_traveler\_argumentation\_profiles\}, Preference profile of the traveler can be one of the following - \{list\_of\_traveler\_preference\_profiles\}, Buying style profile of the traveler can be one of the following - \{list\_of\_traveler\_buyingstyle\_profiles\}, and the Argumentation profile of the travel agent can be one of the following - \{list\_of\_travelagent\_argumentation\_profiles\}. Please give the output as `Traveler Argumentation Profile: \texttt{\textbackslash n} Traveler Preference Profile: \texttt{\textbackslash n} Traveler Buying Style Profile: \texttt{\textbackslash n} Travel Agent Argumentation Profile:'}

\textbf{Input}

\{Conversation\}

\textbf{Output}

Traveler Argumentation Profile: 

Traveler Preference Profile:

Traveler Buying Style Profile:

Travel Agent Argumentation Profile: \\ \hline

\textbf{Dialog Act Prediction Task} \\ \hline
\texttt{\textcolor{purple}{[Task Instruction]}} \texttt{Imagine you are a travel agent negotiating with a potential traveler using an argumentation-based approach. The focus of the negotiation is on a \{\} travel package described as \{\}, where both parties aim to reach a mutually beneficial agreement. Your task is to predict the dialogue act of the target utterance based on the provided dialogue context. Dialog Act may include \{list\_of\_dialog\_acts\}. Please start the output with `Dialog Act Label:'}

\textbf{Input}

\{Dialogue Context\}

\{Target Utterance\}

\textbf{Output}

Dialog Act Label: \\ \hline

\textbf{Response Generation Task} \\ \hline
\texttt{\textcolor{purple}{[Task Instruction]}} \texttt{Imagine you are a travel agent negotiating with a potential traveler using an argumentation-based approach. The focus of the negotiation is on a {} travel package described as {}, where both parties aim to reach a mutually beneficial agreement. Your role as the travel agent is to present compelling arguments, address the traveler’s concerns, and strategically persuade them while maintaining professionalism and flexibility. Given the dialogue context, your task is to generate a coherent and contextually relevant response. Please begin your response with `Response:'}

\textbf{Input}

\{Dialogue Context\}

\textbf{Output}

Response: 

\\\hline

\end{tabular}
\end{adjustbox}
\caption{Prompt template used for GPT-4.1-mini for different tasks. Variables enclosed in curly brackets {} represent placeholders that are filled when the template is instantiated.}
\label{prompt_gpt4_1}
\end{table*}

\begin{table*}[hbt!]
\centering
\begin{adjustbox}{max width=\linewidth}
\begin{tabular}{l|p{19cm}}
\hline

\textbf{Dialogue Context} & [...] \textit{\textbf{Traveler}}: I'm a little concerned about the price of \$43,546.96. Could you break down the cost a bit more for me? I'm not sure if that's justifiable for the package as it stands now. Are there any options for adjusting the itinerary or accommodations to bring the price down? \\
 & \textit{\textbf{Agent}}: I appreciate you're looking for a great deal, and \$37,203.88 is certainly a tempting offer. However, to give you the best possible experience and ensure we can secure all the elements of the package you've requested, we need to consider a price closer to \$41,277.89. This will allow us to give you the best hotel and  include a unique experience, secure premium flights for high-quality experience you deserve. Would you be open to discussing that? \\
 & \textit{\textbf{Traveler}}: I appreciate the package you've put together, but \$41,277.89 is a bit more than I'm comfortable with. Would you be willing to work with me on a price closer to \$37,203.88? \\ \hdashline
 \textbf{Ground-truth} & \textbf{\textit{Agent}}: This package is carefully crafted to offer the best possible value, and to ensure we can deliver on all the amazing experiences included, we've arrived at a price of \$41,277.89.  While I understand you might be looking for a lower price, this figure represents the absolute minimum we can offer to maintain the high quality and comprehensive nature of the tour. \\ \hdashline
\textbf{Llama-2-chat (FT-\textit{Vanilla})} &  Okay, so for this incredible tour package, with everything included, the total cost comes to \$41,277.89. I understand that's a significant investment, but trust me, it's truly worth it. This price includes all flights, accommodations, activities, and even some meals, so you won't have to worry about any hidden costs. I've carefully crafted this itinerary to give you the best possible experience, and I'm confident that it's a fantastic value.  While I can't go lower than this price, I'm happy to discuss any potential adjustments to the package if you have any specific areas you'd like to change.\\\hdashline
\textbf{Vicuna (FT-\textit{Vanilla})} & This package has been carefully designed to provide you with the best possible experience, and after considering everything included, the price comes to \$41,277.89. I understand that you may be seeking a lower price, but this is the absolute minimum we can offer to maintain the high-quality experiences and comprehensive nature of the tour. I’m confident that this reflects the value you’ll receive. \\\hdashline

\textbf{Llama-2-chat (FT-\textit{Ours})} & I understand your budget concerns and appreciate your flexibility. We can explore adjusting accommodations, such as switching some premium hotel nights to a boutique stay or tweaking the itinerary while maintaining key experiences. Would you be open to discussing specific preferences to better align with your target of \$37,203.88? \\\hdashline

\textbf{Vicuna (FT-\textit{Ours})} & I appreciate your concerns, and I want to ensure you get the best value while maintaining the quality of your experience. The price of \$41,277.89 represents the minimum necessary to provide the high-quality accommodations, curated experiences, and seamless logistics that make this trip exceptional. While I understand your goal of adjusting the cost, reducing it significantly would impact key elements of the package. That said, I’d be happy to explore reasonable modifications to tailor it further, perhaps adjusting certain accommodations or activities. Would you like me to walk you through some potential refinements? \\ \hline 

\textbf{Dialogue Context} & [...] \textit{\textbf{Traveler}}: Hi, I'm looking at the tour package and I'm not really interested in the photography tour. Could we remove that from the package and adjust the price accordingly?\\
 & \textit{\textbf{Agent}}: Okay, so if we remove the photography tour from the package, the total cost will come down to \$37,817.13. Does that sound better for you? \\
 & \textit{\textbf{Traveler}}: I'm really interested in this package, but my budget is a bit tighter than that. Would you be willing to work with me on a price closer to \$37,149.59? I'm sure we can find a mutually agreeable solution.\\ \hdashline
 \textbf{Ground-truth} & \textbf{\textit{Agent}}: Okay, so the total cost for this amazing tour package comes out to \$37,817.13. That price includes everything, from flights and accommodations to all the incredible excursions and activities we've discussed.  I understand that's a significant investment, but I can assure you it's a truly exceptional value for the experience you'll receive. Now, I'm very confident in the value of this package, and I'm not able to go below that price point.  However, I'm happy to explore any other options with you, such as potentially adjusting the itinerary to see if we can find some savings.  What are your thoughts? \\ \hdashline
\textbf{Llama-2-chat (FT-\textit{Vanilla})} & I appreciate you taking the time to put together that budget, and I see you're looking for a great deal. While \$37,149.59 is a fantastic price point, I want to make sure you get the best value for your trip.  Considering the premium experiences and amenities included in this package, I'm proposing a slightly adjusted price of \$42,231.07. This allows us to secure the top-rated accommodations, exclusive excursions, and personalized service you deserve.  What do you think? \\ \hdashline
\textbf{Vicuna (FT-\textit{Vanilla})} & I understand you're working within a specific budget, and I want to make sure you get the best value for your trip. Based on the current package, the total price is \$37,817.13, which includes everything we've discussed—flights, accommodations, and all the exciting excursions. I'm confident that this price reflects the quality of the experiences you’ll have, and I’m unable to lower it further. However, I’d be happy to discuss options for adjusting the itinerary if you'd like to explore some alternatives to bring the cost closer to your budget. What do you think? \\ \hdashline
\textbf{Llama-2-chat (FT-\textit{Ours})} & I understand your budget and appreciate your interest. The \$37,817.13 covers flights, accommodations, and excursions. While I can't lower it further, we can adjust the itinerary to better match your \$37,149.59 target. Would you like to explore options?  \\ \hdashline

\textbf{Vicuna (FT-\textit{Ours})} & I appreciate your offer and understand the importance of staying within budget. The \$37,817.13 package ensures a high-quality experience, covering all key inclusions such as flights, premium stays, and curated excursions. While this is the best price we can offer for the current package, I’m happy to explore small itinerary adjustments to bring it closer to your \$37,149.59 target. Let’s review which aspects matter most to you, would you like to discuss possible tweaks? \\\hline

\end{tabular}
\end{adjustbox}
\caption{Example responses generated by different LLMs under different task setup. [...] represents the truncated context for simplicity.}
\label{qual_analysis}
\end{table*}

\begin{table*}[!hbt]\footnotesize
\centering
\begin{adjustbox}{max width=\linewidth}
\begin{tabular}{p{2cm} p{10cm}}
\hline
\multicolumn{2}{c}{\textbf{PR: 5, BE: 4, OF: 4, F: 4, E: 3, C: 4, Faith.: Yes}} \\ \hline
\textbf{Speaker} & \textbf{Dialogue} \\
\hline
Travel Agent & Hi! I'm here to help plan your ideal trip. Are you looking for adventure, relaxation, a city break, or something else? \\
Traveler & I want a package packed with attractions-landmarks, events, scenic views. I love history, culture, and exciting experiences. \\
Travel Agent & The Sight Seeker package might be perfect. It’s \$54,007 and includes a cottage stay, fine dining, private transport, VR experiences, mindfulness sessions, and personal assistance. Interested? \\
Traveler & VR sounds great! What kind of experiences are included? Anything cultural or historical? \\
Travel Agent & Yes! You can explore a Marrakech market or dive into the Great Barrier Reef-virtually. It’s immersive and memorable. \\
Traveler & I love it, but \$54,000 is too much. Could you consider \$42,658? \\
Travel Agent & This price includes top-tier amenities and experiences. It’s tough to go lower while maintaining quality. \\
Traveler & Okay, how about \$45,564.59? Can we make that work? \\
Travel Agent & Absolutely! This package suits you well. Let’s confirm your details and book it. Thank you! \\
\hline
\end{tabular}
\end{adjustbox}
\caption{An interaction between the traveler and travel agent (best-performing model) engaged in argumentation-based negotiation.}
\label{casestudy_multiturn}
\end{table*}

\end{document}